\documentclass{article} 
\usepackage{iclr2023_conference}
\usepackage{times}
\usepackage{natbib}

\usepackage{amsmath,amsfonts,bm}

\def\eqref#1{equation~\ref{#1}}

\def\1{\bm{1}}

\DeclareMathAlphabet{\mathsfit}{\encodingdefault}{\sfdefault}{m}{sl}
\SetMathAlphabet{\mathsfit}{bold}{\encodingdefault}{\sfdefault}{bx}{n}

\newcommand{\E}{\mathop{\mathbb{E}}}

\DeclareMathOperator*{\argmax}{arg\,max}

\usepackage{hyperref}
\usepackage{xcolor}
\hypersetup{
    colorlinks,
    linkcolor={red!50!black},
    citecolor={blue!50!black},
    urlcolor={blue!80!black}
}
\usepackage{url}
\usepackage{graphicx, wrapfig}
\usepackage{bm}
\usepackage{subfigure}
\usepackage{tikz}
\usepackage{cancel, amsmath}
\usepackage{breqn}
\usetikzlibrary{bayesnet}
\usepackage{algorithm,algpseudocode}

\usepackage{enumitem}

\title{Learning Latent Structural Causal Models}

\author{Jithendaraa Subramanian\thanks{Correspondence to jithen.subra@gmail.com} \\
Mila, McGill University\\ 
\And
Yashas Annadani \\
KTH Stockholm \\
\And
Ivaxi Sheth \\
Mila, ÉTS Montréal \\
\And
Nan Rosemary Ke \\
Mila, Deepmind \\
\And
Tristan Deleu \\
Mila, Université de Montréal \\
\And
Stefan Bauer \\
KTH Stockholm \\
\And
Derek Nowrouzezahrai \\
Mila, McGill University \\
\AND
Samira Ebrahimi Kahou \\
Mila, ÉTS Montreal, CIFAR AI Chair \\
}

\preprint
\begin{document}

\maketitle

\begin{abstract}
Causal learning has long concerned itself with the accurate recovery of underlying causal mechanisms. Such causal modelling enables better explanations of out-of-distribution data. Prior works on causal learning assume that the high-level causal variables are given. However, in machine learning tasks, one often operates on low-level data like image pixels or high-dimensional vectors. In such settings, the entire Structural Causal Model (SCM) -- structure, parameters, \textit{and} high-level causal variables -- is unobserved and needs to be learnt from low-level data. We treat this problem as Bayesian inference of the latent SCM, given low-level data. For linear Gaussian additive noise SCMs, we present a tractable approximate inference method which performs joint inference over the causal variables, structure and parameters of the latent SCM from random, known interventions. Experiments are performed on synthetic datasets and a causally generated image dataset to demonstrate the efficacy of our approach. We also perform image generation from unseen interventions, thereby verifying out of distribution generalization for the proposed causal model.
\end{abstract}


\section{Introduction}
Learning variables of interest and uncovering causal dependencies is crucial for intelligent systems to reason and predict in scenarios that differ from the training distribution. In the causality literature, causal variables and mechanisms are often assumed to be known. This knowledge enables reasoning and prediction under unseen interventions. In machine learning, however, one does not have direct access to the underlying variables of interest nor the causal structure and mechanisms corresponding to them. Rather, these have to be learned from observed low-level data like pixels of an image which are usually high-dimensional. Having a learned causal model can then be useful for generalizing to out-of-distribution data~\citep{nino_ood, ke2021systematic}, estimating the effect of interventions \citep{pearl_2009, tcrl}, disentangling underlying factors of variation \citep{replearn, causal_desiderata}, and transfer learning \citep{ocacl, bengio2019meta}.

\begin{figure}[t] 
    \centering
    \includegraphics[width=12cm]{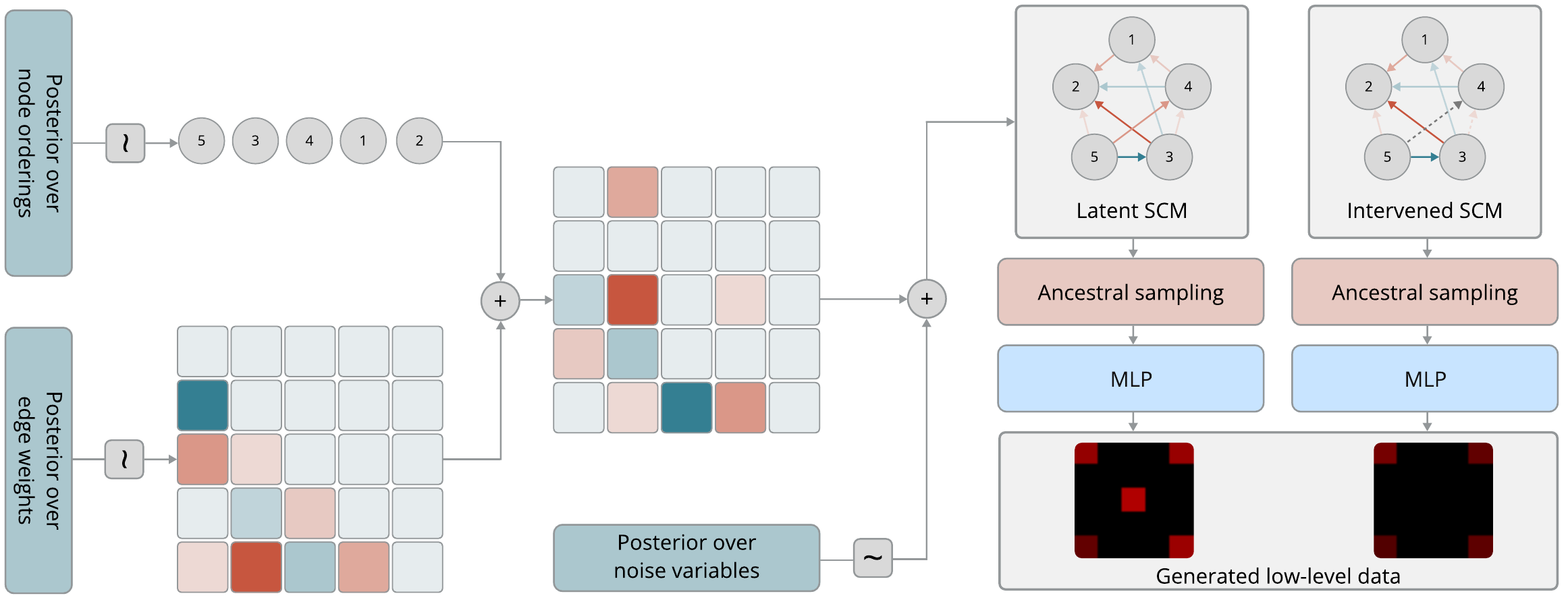}
    \caption{Model architecture of the proposed generative model for the Bayesian latent causal discovery task to learn latent SCM from low-level data.}
    \label{overview}
\end{figure}

Structure learning \citep{pc, zheng2018dags} learns the structure and parameters of the Structural Causal Model (SCM) \citep{pearl_2009} that best explains some observed high-level causal variables. In causal machine learning and representation learning, however, these causal variables may no longer be observable. This serves as the motivation for our work. We address the problem of learning the entire SCM -- consisting its causal variables, structure and parameters -- which is latent, by learning to generate observed low-level data. Since one often operates in low-data regimes or non-identifiable settings, we adopt a Bayesian formulation so as to quantify epistemic uncertainty over the learned latent SCM. Given a dataset, we use variational inference to learn a joint posterior over the causal variables, structure and parameters of the latent SCM. To the best of our knowledge, ours is the first work to address the problem of causal discovery in linear Gaussian latent SCMs from low-level data, where causal variables are unobserved. Our contributions are as follows:

\begin{itemize}
    \item We \emph{propose a general algorithm for Bayesian causal discovery in the latent space of a generative model}, learning a distribution over causal variables, structure and parameters in linear Gaussian latent SCMs with random, known interventions. Figure~\ref{overview} illustrates an overview of the proposed method.
    
    \item By learning the structure and parameters of a latent SCM, we implicitly induce a joint distribution over the causal variables. Hence, sampling from this distribution is equivalent to ancestral sampling through the latent SCM. As such, \emph{we address a challenging, simultaneous optimization problem} that is often encountered during causal discovery in latent space: one cannot find the right graph without the right causal variables, and vice versa.
    
    \item On a synthetically generated dataset and an image dataset used to benchmark causal model performance \citep{ke2021systematic}, we evaluate our method along three axes -- uncovering causal variables, structure, and parameters -- consistently outperforming baselines. We demonstrate its ability to perform image generation from unseen interventional distributions.
\end{itemize}

\section{Preliminaries}

\subsection{Structural Causal Models}
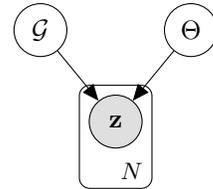
\begin{wrapfigure}{r}{4cm} 
    \centering
    \vspace{-15pt}
    \begin{tikzpicture}
        \node[obs]                                 (z) {$\mathbf{z}$}; 
        \node[latent, above=of z, xshift=-1cm, yshift=-0.5cm] (g) {${\mathcal{G}}$}; 
        \node[latent, above=of z, xshift=1cm, yshift=-0.5cm]  (t) {$\Theta$};
        \edge {g,t} {z}; 
        \plate {z} {(z)} {$N$};  
    \end{tikzpicture}
    \caption{BN for prior works in causal discovery and structure learning}
    \vspace{-10pt}
    \label{fig2}
\end{wrapfigure}

A Structural Causal Model (SCM) is defined by a set of equations which represent the mechanisms by which each endogenous variable $z_i$ depends on its direct causes $z^{\mathcal{G}}_{pa(i)}$ and a corresponding exogenous noise variable $\epsilon_i$. The direct causes are subsets of other endogenous variables. If the causal parent assignment is assumed to be acyclic, then an SCM is associated with a Directed Acyclic Graph (DAG) $\mathcal{G}=(V,E)$, where V corresponds to the endogenous variables and $E$ encodes direct cause-effect relationships. The exact value taken on by a causal variable $z_i$, is given by local causal mechanisms $f_i$ conditional on $z_{pa(i)}^{{\mathcal{G}}}$, the parameters $\Theta_i$, and the node's noise variable $\epsilon_i$, as given in equation \ref{eq1}. For linear Gaussian additive noise SCMs with equal noise variance, i.e., the setting that we focus on in this work, all $f_i$'s are linear functions, and $\Theta$ denotes the weighted adjacency matrix $W$, where each $W_{ji}$ is the edge weight from $j \rightarrow i$. The linear Gaussian additive noise SCM thus reduces to equation \ref{eq2}, 
\begin{center}
\begin{tabular}{p{5cm}p{6cm}}
    \begin{equation} \label{eq1}
        z_i = f_i(z_{pa(i)}^{{\mathcal{G}}}, \Theta, \epsilon_i)~,
    \end{equation}
  &
    \begin{equation} \label{eq2}
        z_i = \sum_{j \in pa_{{\mathcal{G}}}(i)}{W_{ji} \cdot z_j}  + \epsilon_i~.
    \end{equation}
\end{tabular}
\end{center}

\subsection{Causal Discovery}

Structure learning in prior work refers to learning a DAG according to some optimization criterion with or without the notion of causality (e.g., \cite{graphvae}). The task of causal discovery on the other hand, is more specific in that it refers to learning the structure (also parameters, in some cases) of SCMs, and subscribes to causality and interventions like that of~\citet{pearl_2009}. That is, the methods aim to estimate $({\mathcal{G}}, \Theta)$. These approaches often resort to modular likelihood scores over causal variables -- like the BGe score \citep{Geiger1994LearningGN, kuipers2022efficient} and BDe score \citep{heckerman1995learning} -- to learn the right structure. However, these methods all assume a dataset of observed causal variables. These approaches either obtain a maximum likelihood estimate,
\begin{equation} \label{eq3}
    {\mathcal{G^*}} = \argmax_{  {\mathcal{G}} } p(Z \mid {\mathcal{G}}) \;\;\;  \text{or} \;\;\; ({\mathcal{G^*}}, \Theta^*) = \argmax_{  {\mathcal{G}}, \Theta } p(Z \mid {\mathcal{G}}, \Theta) ~,
\end{equation}
or in the case of Bayesian causal discovery~\citep{heckerman1997bayesian}, variational inference is typically used to approximate a joint posterior distribution $q_{\phi}({\mathcal{G}}, \Theta)$ to the true posterior $p({\mathcal{G}}, \Theta \mid Z)$ by minimizing the KL divergence between the two,

\begin{equation} \label{eq4}
       D_{\text{KL}}( q_{\phi}({\mathcal{G}}, \Theta)\, ||\, p({\mathcal{G}}, \Theta \mid Z)) = - \mathbb{E}_{({\mathcal{G}}, \Theta) \sim q_{\phi}} \biggl[ \log p(Z \mid {\mathcal{G}}, \Theta) - \log \frac{q_{\phi}({\mathcal{G}}, \Theta)}{p({\mathcal{G}}, \Theta)} \biggr]~,
\end{equation}

where $p({\mathcal{G}}, \Theta)$ is a prior over the structure and parameters of the SCM -- possibly encoding DAGness, sparse connections, or low-magnitude edge weights. Figure \ref{fig2} shows the Bayesian Network (BN) over which inference is performed for causal discovery tasks.

\subsection{Latent causal discovery}
\begin{wrapfigure}{r}{5cm} 
\centering
    \vspace{-15pt}
    \begin{tikzpicture} 
        \node[obs]                                 (x) {$\mathbf{x}$};
        \node[latent, above=of x, yshift=-0.4cm]   (z) {$\mathbf{z}$}; 
        \node[latent, above=of z, xshift=-1cm, yshift=-0.7cm] (g) {${\mathcal{G}}$}; 
        \node[latent, above=of z, xshift=1cm, yshift=-0.7cm]  (t) {$\Theta$};
        \edge {g,t} {z}; 
        \edge {z} {x};
        \plate {xz} {(x)(z)} {$N$} ;
    \end{tikzpicture}
    \caption{BN for the latent causal discovery task that generalizes standard causal discovery setups}
    \vspace{-10pt}
    \label{fig3}
\end{wrapfigure}
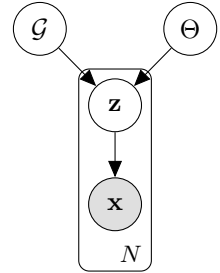

In more realistic scenarios, the learner does not directly observe causal variables and they must be learned from low-level data. The causal variables, structure, and parameters are part of a latent SCM. The goal of causal representation learning models is to perform inference of, and generation from, the true latent SCM. \citet{causalvae} proposes a Causal VAE but is in a supervised setup where one has labels on causal variables and the focus is on disentanglement. \citet{causalgan} present causal generative models trained in an adversarial manner but assumes observations of causal variables. Given the right causal structure as a prior, the work focuses on generation from conditional and interventional distributions. 

In both the causal representation learning and causal generative model scenarios mentioned above, the Ground Truth (GT) causal graph and parameters of the latent SCM are arbitrarily defined on real datasets and the setting is supervised. Contrary to this, our setting is unsupervised and we are interested in recovering the GT underlying SCM and causal variables that generate the low-level observed data -- we define this as the problem of \emph{latent causal discovery}, and the BN over which we want to perform inference on is given in figure \ref{fig3}. In the upcoming sections, we discuss related work, formulate our problem setup and propose an algorithm for Bayesian latent causal discovery, evaluate with experiments on causally created vector data and image data, and perform sampling from unseen interventional image distributions to showcase generalization of learned latent SCMs.

\section{Related Work} 
Prior work can be classified into Bayesian \citep{koivisto2004exact, bacd, friedman2013being} or maximum likelihood \citep{DCDI, nofears, ng2022masked} methods, that learn the structure and parameters of SCMs using either score-based \citep{10.2307/2291091, 720554, heckerman1995learning} or constraint-based \citep{CHENG200243, lehmann2005testing} approaches.

\textbf{Causal discovery and structure learning:} Work in this category assume causal variables are observed and do not operate on low-level data~\citep{pc, gadget, dagnocurl, https://doi.org/10.48550/arxiv.2208.14571}. \citet{gdseev} prove identifiability of linear Gaussian SCMs with equal noise variances. \citet{bengio2019meta} use the speed of adaptation as a signal to learn the causal direction. \citet{ke2019learning} explore learning causal models from unknown, while \citet{scherrer2021learning, tigas2022interventions, abcd, toth2022active} focus on active learning and experimental design setups on how to perform interventions to efficiently learn causal models. Transformer~\citep{vaswani2017attention} based approach learns structure from synthetic datasets and generalize to naturalistic graphs~\citep{ke2022learning}. \citet{zheng2018dags} introduce an acyclicity constraint that penalizes cyclic graphs, thereby restricting search close to the DAG space. \citet{grandag} leverages this constraint to learn DAGs in nonlinear SCMs. \citet{dynotears, citris} perform structure learning on temporal data. 

\textbf{Latent variable models with predefined structure}: Examples include VAE \citep{vae, vae_rezende} which has an independence assumption between latent variables. To overcome this, \cite{laddervae} and \cite{zhao} define latent variables with a chain structure in VAEs. \cite{vae_iaf} uses inverse autoregressive flows to improve upon the diagonal covariance of latent variables in VAEs.

\textbf{Latent variable models with learned structure}: GraphVAE \citep{graphvae} learns the edges between latent variables without incorporating notions of causality. \citet{brehmer2022weakly} present identifiability theory for learning causal representations and propose a practical algorithm for this under the assumption of having access to pairs of observational and interventional data.

\textbf{Supervised causal representation learning}:
\citet{causalgan, dear, can} introduce generative models that use an SCM-based prior in latent space. In \cite{dear}, the goal is to learn causally disentangled variables. \cite{causalvae} learn a DAG on CelebA and a causally generated pendulum image dataset but assume complete access to the causal variables. \cite{csii} establishes observable causal footprints in images.

\section{Learning latent SCMs from low-level data}
\subsection{Problem Scenario}

We are given a dataset $\mathcal{D} = \{\mathbf{x}^{(1)},...,\mathbf{x}^{(N)}\}$, where each $\mathbf{x}^{(i)}$ is a high-dimensional observed data -- for simplicity, we assume $\mathbf{x}^{(i)}$ is a vector in $\mathbb{R}^{\mathrm{D}}$ but the setup extends to other inputs as well (like images, as we will see in the next section). We assume that there exist latent variables $\mathbf{Z}=\{\mathbf{z}^{(i)} \in \mathbb{R}^d\}_{i=1}^{N}$ with $d \leq \mathrm{D}$, that explain the data, and these latent variables have an SCM with structure $\mathcal{G}_{GT}$ and parameters $\Theta_{GT}$ associated with them. 
We wish to invert the data generation process $g:(\mathbf{Z}, \mathcal{G}, \Theta) \rightarrow \mathcal{D}$ where the causal variables $\mathbf{Z}$ are in the latent space. In the setting, we also have access to the intervention targets $\mathcal{I} = \{ \mathcal{I}^{(i)} \}_{i=1}^N$ where each $\mathcal{I}^{(i)} \in \{0, 1\}^d$. The $j^{\text{th}}$ dimension of $\mathcal{I}^{(i)}$ takes a value of 1 if node $j$ was intervened on in data sample $i$, and 0 otherwise.

\subsection{General method}
We aim to obtain a posterior estimate over the entire latent SCM, $p(\mathbf{Z}, {\mathcal{G}}, \Theta \mid \mathcal{D})$. Computing the true posterior analytically requires calculating the marginal likelihood $p(\mathcal{D})$ which gets quickly intractable due to the number of possible DAGs growing super-exponentially with respect to the number of nodes. Thus, we resort to variational inference~\citep{blei2017variational} that provides a tractable way to learn an approximate posterior $q_{\phi}(\mathbf{Z}, {\mathcal{G}}, \Theta)$ with variational parameters $\phi$, close to the true posterior $p(\mathbf{Z}, {\mathcal{G}}, \Theta \mid \mathcal{D})$ by maximizing the Evidence Lower Bound (ELBO),
\begin{equation} \label{eq5}
\mathcal{L}(\psi, \phi) = \E_{q_{\phi}(\mathbf{Z}, {\mathcal{G}}, \Theta)} \biggl[ \log p_\psi(\mathcal{D} \mid \mathbf{Z}, {\mathcal{G}}, \Theta) - \log  \frac{ q_{\phi}(\mathbf{Z}, {\mathcal{G}}, \Theta) }{p(\mathbf{Z}, {\mathcal{G}}, \Theta)} \biggr]~,
\end{equation}

where $p(\mathbf{Z}, {\mathcal{G}}, \Theta)$ is the prior, $p_\psi(\mathcal{D} \mid \mathbf{Z}, {\mathcal{G}}, \Theta)$ is the likelihood model with parameters $\psi$, the likelihood model maps the latent variables $\mathbf{Z}$ to high-dimensional vectors $\mathbf{X}$. An approach to learn this posterior could be to factorize it as
\begin{equation} \label{eq6}
    q_{\phi}(\mathbf{Z}, {\mathcal{G}}, \Theta) = q_{\phi}(\mathbf{Z}) \cdot q_{\phi}({\mathcal{G}}, \Theta \mid \mathbf{Z})~.
\end{equation}

Given a way to obtain $q_{\phi}(\mathbf{Z})$, the conditional $q_{\phi}({\mathcal{G}}, \Theta \mid \mathbf{Z})$ can be obtained using existing Bayesian structure learning methods. Otherwise, one has to perform a hard simultaneous optimization which would require alternating optimizations on $\mathbf{Z}$ and on $(\mathcal{G}, \Theta)$ given an estimate of $\mathbf{Z}$. Difficulty of such an alternate optimization is discussed in \citet{brehmer2022weakly}. 

\textbf{Alternate factorization of the posterior}: Rather than factorizing as in equation \ref{eq6}, we propose to factorize according to the BN in figure \ref{fig3}. This is given by $q_{\phi}(\mathbf{Z}, {\mathcal{G}}, \Theta) = q_{\phi}(\mathbf{Z} \mid \mathcal{G}, \Theta) \cdot q_{\phi}({\mathcal{G}}, \Theta)$. The advantage of this factorization is that the distribution over $\mathbf{Z}$ is completely determined from the SCM given $(\mathcal{G}, \Theta)$ and exogenous noise variables (assumed to be Gaussian). Thus, the prior $p(\mathbf{Z}\mid \mathcal{G}, \Theta)$ and the posterior $p(\mathbf{Z}\mid \mathcal{G}, \Theta, \mathcal{D}) = q_{\phi}(\mathbf{Z} \mid \mathcal{G}, \Theta)$ are identical. This conveniently avoids the hard simultaneous optimization problem mentioned above since optimizing for $q_\phi(\mathbf{Z})$ is not necessary. Equation \ref{eq5} can then be simplified as 
\begin{equation} \label{eq7}
    \mathcal{L}(\psi, \phi)=\E_{q_{\phi}(\mathbf{Z}, \mathcal{G}, \Theta)} \biggl[ \log p_\psi(\mathcal{D} \mid \mathbf{Z}) - \log  \frac{ q_{\phi}(\mathcal{G}, \Theta) }{p(\mathcal{G}, \Theta)} - \cancelto{0}{\log  \frac{ q_{\phi}(\mathbf{Z} \mid \mathcal{G}, \Theta) }{p(\mathbf{Z} \mid \mathcal{G}, \Theta)}} \biggr] ~.
\end{equation}
Such a posterior can be used to obtain an SCM by sampling $\mathcal{G}$ and $\Theta$ from the approximated posterior. As long as the samples $\mathcal{G}$ are always acyclic, one can perform ancestral sampling through the SCM to obtain samples corresponding to the causal variables $\hat{\mathbf{z}}^{(i)}$. For additive noise models like in equation~\ref{eq2}, these samples are already reparameterized and differentiable with respect to their parameters. The samples of causal variables are then fed to the likelihood model to predict $\hat{\mathbf{x}}^{(i)}$ to reconstruct the observed data $\mathbf{x}^{(i)}$.

\subsection{Posterior parameterizations and priors} 
For linear Gaussian latent SCMs, which is the focus of this work, learning a posterior over $(\mathcal{G}, \Theta)$ is equivalent to learning $q_{\phi}(W, \Sigma)$ -- a posterior over weighted adjacency matrices $W$ and noise covariances $\Sigma$. We follow an approach similar to \citep{bcdnets}. We express $W$ via a permutation matrix $P$\footnote{A permutation matrix $P\in \{0,1\}^{d\times d}$ is a bistochastic  matrix with $\sum_i p_{ij} = 1 \forall j$ and $\sum_j p_{ij}=1 \forall i$.} and a lower triangular edge weight matrix $L$, according to $W = P^T L^T P$. Here, $L$ is defined in the space of all weighted adjacency matrices with a fixed node ordering where node $j$ can be a parent of node $i$ only if $j > i$. Search over permutations corresponds to search over different node orderings and thus, $W$ and $\Sigma$ parameterize the space of SCMs. Further, we factorize the approximate posterior $q_{\phi}(P, L, \Sigma)$ as
\begin{equation} \label{eq8}
    q_{\phi}{(G, \Theta)} \equiv q_{\phi}{(W, \Sigma)} \equiv q_{\phi}{(P, L, \Sigma)} = q_{\phi}{(P \mid L, \Sigma)} \cdot q_{\phi}{(L, \Sigma)}~.
\end{equation}

Combining equation \ref{eq7} and \ref{eq8} leads to the following ELBO which has to be maximized (derived in \ref{elbo_derive}), and the overall algorithm for Bayesian latent causal discovery is summarized in algorithm \ref{lbcd_algo},
\begin{align*} \label{elbo_eqn}
    \mathcal{L}(\psi, \phi) \!=\!\!\!\!\!\E_{ q_{\phi}(L, \Sigma)} &
     \Biggl[ \E_{q_{\phi}(P \mid L, \Sigma)} 
     \biggl[ \E_{q_{\phi}(\mathbf{Z} \mid P, L, \Sigma)} 
     [\, \log p_{\psi}(\mathcal{D} \mid \mathbf{Z}) \, ] \!-\! \log  \frac{ q_{\phi}(P \!\mid\! L, \Sigma) }{p(P)} \biggr] 
     \!-\! \log  \frac{ q_{\phi}(L, \Sigma) }{p(L)p(\Sigma)} \Biggr]  \tag{9
     }.
\end{align*}

\textbf{Distribution over $(L, \Sigma)$}: The posterior distribution $q_{\phi}{(L, \Sigma)}$ has $\bigl( \frac{d(d-1)}{2} + 1 \bigr)$ elements to be learnt in the equal noise variance setting. This is parameterized as a diagonal covariance normal distribution. For the prior $p(L)$ over the edge weights, we promote sparse DAGs by using a horseshoe prior \citep{horseshoe}, similar to \cite{bcdnets}. A Gaussian prior is defined over $\log \Sigma$.

\begin{algorithm}[ht]
\caption{Bayesian latent causal discovery to learn $\mathcal{G}$, $\Theta$, $\mathbf{Z}$ from high dimensional data} \label{lbcd_algo}
\begin{algorithmic}[1]
\Require{$\mathcal{D}, \mathcal{I}$}
\Ensure{$\mathcal{G}$, $\Theta$, $\mathbf{Z}$}
\State Initialize $q_{\phi}(L, \Sigma)$, $\text{MLP}_{\phi(T)}$, $p_\psi(\mathbf{X} \mid \mathbf{Z})$, $\tau$ and set learning rate $\alpha$  
\For{$\text{num\_epochs}$}
    \State{$(\widehat{L}, \widehat{\Sigma}) \sim q_{\phi}(L, \Sigma)$}
    \State{$T \leftarrow \text{MLP}_{\phi(T)}(\widehat{L}, \widehat{\Sigma})$}\Comment{Compute logits for sampling from $q_{\phi}(P \mid L, \Sigma)$}
    \State{$\gamma \in \mathbb{R}^{d \times d} \sim$ standard Gumbel}
    \State{$\widehat{P}_{\text{soft}} \leftarrow Sinkhorn((T + \gamma)/ \tau)$} \State{$\widehat{P}_{\text{hard}} \leftarrow Hungarian(\widehat{P}_{\text{soft}})$}
    \State{$\widehat{W} \leftarrow \widehat{P}^T \widehat{L}^T \widehat{P}$}
    \For{$i \gets 1$ to $N$}
        \State{$\mathcal{C}^{(i)} \leftarrow \text{argwhere}(\mathcal{I}^{(i)}= 1) $}
        \State{$\widetilde{W} = \text{copy}(\widehat{W})$}
        \State{$\widetilde{W}[:, \mathcal{C}^{(i)}] \leftarrow 0$}\Comment{Mutated weighted adjacency matrix according to $\mathcal{I}^{(i)}$}
        \State{$\widehat{W}_{\mathcal{I}^{(i)}} \leftarrow \widetilde{W}$}
        \State{$\hat{\mathbf{z}}^{(i)} \leftarrow \text{AncestralSample}(\widehat{W}_{\mathcal{I}^{(i)}}, \widehat{\Sigma})$}
    \EndFor
    \State{$\hat{\mathbf{Z}} \leftarrow \{ \hat{\mathbf{z}}^{(i)} \}_{i=1}^N$ }
    \State{$\hat{\mathcal{D}} \sim p_\psi(\mathbf{X} \mid \mathbf{Z})$}
    \State{$\psi \leftarrow \psi + \alpha \cdot \nabla_{\psi}(\mathcal{L}(\psi, \phi))$}\Comment{Update network parameters}
    \State{$\phi \leftarrow \phi + \alpha \cdot \nabla_{\phi}(\mathcal{L}(\psi, \phi))$}
\EndFor
\State \Return {$\text{binary}(\widehat{W}), (\widehat{W}, \widehat{\Sigma}), \mathbf{\hat{Z}}$}
\end{algorithmic}
\end{algorithm}

\textbf{Distribution over $P$}: Since the values of $P$ are discrete, performing a discrete optimization is combinatorial and becomes quickly intractable with increasing $d$. This can be handled by relaxing the discrete permutation learning problem to a continuous optimization problem. This is commonly done by introducing a Gumbel-Sinkhorn \citep{gs_perm_learning} distribution and where one has to calculate $S((T + \gamma)/\tau)$, where $T$ is the parameter of the Gumbel-Sinkhorn, $\gamma$ is a matrix of standard Gumbel noise, and $\tau$ is a temperature parameter. The logits $T$ are predicted by passing the predicted $(L, \Sigma)$ through an MLP. In the limit of infinite iterations and as $\tau \to 0$, sampling from the distribution returns a doubly stochastic matrix. During the forward pass, a hard permutation $P$ is obtained by using the Hungarian algorithm \citep{hungarian} which allows $\tau \to 0$. During the backward pass, a soft permutation is used to calculate gradients similar to \citep{bcdnets, diff_dag_sampling}. We use a uniform prior $p(P)$ over permutations.  

\section{Experiments and evaluation} \label{experiments}
We perform experiments to evaluate the learned $(\mathbf{Z}, {\mathcal{G}}, {\Theta})$ of the true linear Gaussian latent SCM from high-dimensional data. We aim to highlight the performance of our proposed method on latent causal discovery. As proper evaluation in such a setting would require access to the GT causal graph that generated the high-dimensional observations, we test our method against baselines on synthetically generated vector data and in the realistic case of learning the SCM from pixels in the chemistry environment dataset of~\citep{ke2021systematic}, both of which have a GT causal structure to be compared with. Further, we evaluate the ability of our model to sample images from unseen interventional distributions. 

\textbf{Baselines}: Since we are, to the best of our knowledge, the first to study this setting of learning latent SCMs from low level observations, we are not aware of baseline methods that solve this task. However, we compare our approach against two baselines: (i) Against VAE that has a marginal independence assumption between latent variables and thus have a predefined structure in the latent space, and (ii) against GraphVAE \citep{graphvae} that learns a structure between latent variables. For all baselines, we treat the learned latent variables as causal variables and compare the recovered structure, parameters, and causal variables recovered. Since GraphVAE does not learn the parameters, we fix the edge weight over all predicted edges to be 1.

\begin{figure}[t] 
    \centering
    \includegraphics[width=14cm]{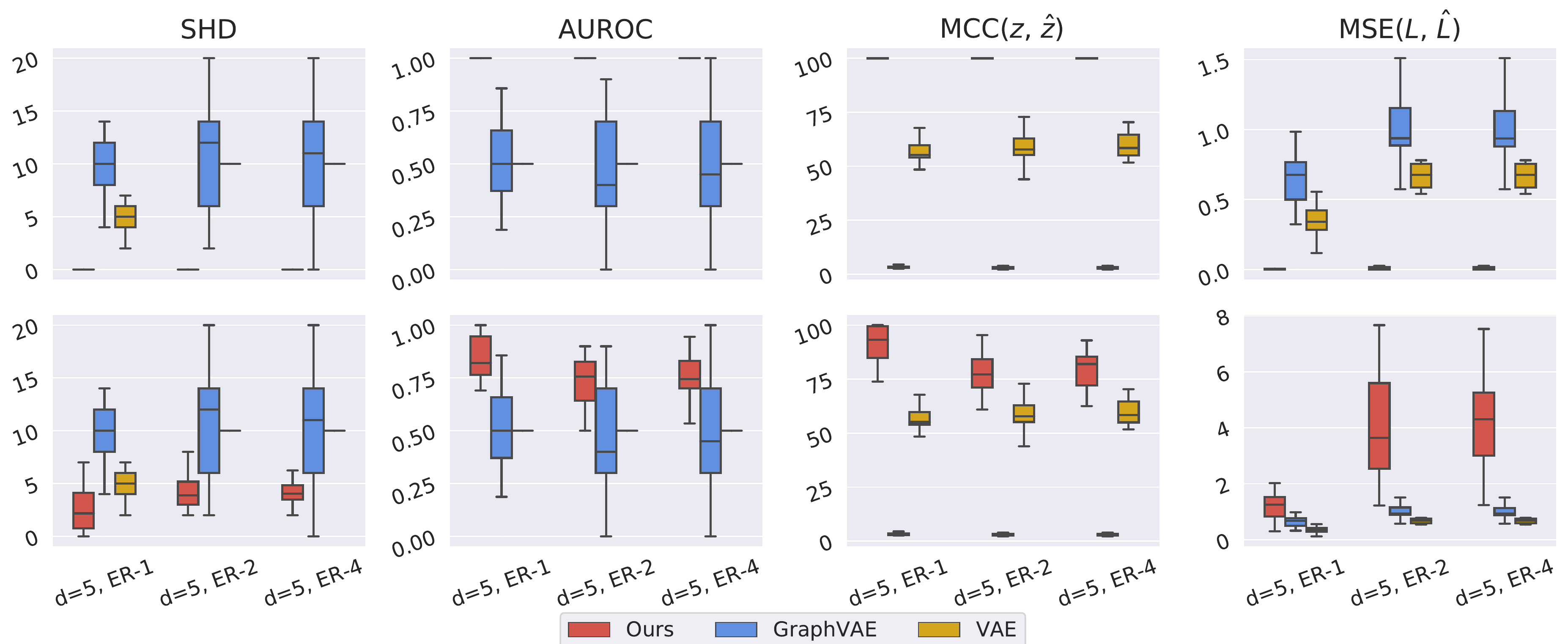}
    \vspace{-0.55cm}
    \caption{Learning the latent SCM (i) given a node ordering (top) and (ii) over node orderings (bottom) for \textbf{linear} projection of causal variables for $d = 5$ nodes, $\mathrm{D} = 100$ dimensions: \textbf{$\mathbb{E}$-SHD ($\downarrow$)}, \textbf{AUROC ($\uparrow$)}, \textbf{MCC ($\uparrow$)}, \textbf{MSE ($\downarrow$)}.}
    \label{d5_lineardbcd}
\end{figure}

\textbf{Evaluation metrics}: \textit{To evaluate the learned structure}, we use two metrics commonly used in the literature -- the expected Structural Hamming Distance (\textbf{$\mathbb{E}$-SHD}, lower is better) obtains the SHD (number of edge flips, removals, or additions) between the predicted and GT graph and then takes an expectation over SHDs of posterior DAG samples, and the Area Under the Receiver Operating Characteristic curve (\textbf{AUROC}, higher is better) where a score of 0.5 corresponds to a random DAG baseline. \textit{To evaluate the learned parameters} of the linear Gaussian latent SCM, we use the Mean Squared Error (\textbf{MSE}, lower is better) between the true and predicted edge weights. \textit{To evaluate the learned causal variables}, we use the Mean Correlation Coefficient (\textbf{MCC}, higher is better) following \cite{pmlr-v54-hyvarinen17a, mcc2} and \cite{ https://doi.org/10.48550/arxiv.2204.04606} which calculates a score between the true and predicted causal variables. See appendix \ref{train_curves} and \ref{detail_eval} for training curves and more extensive evaluations of the experiments along other metrics. Our results are presented over 20 random DAGs with a learning rate of $0.0008$. All our implementations are in JAX \citep{jax2018github}.

\subsection{Experiments on Synthetic Data}
We evaluate our proposed method with the baselines on synthetically generated dataset, where we have complete control over the data generation procedure.

\subsubsection{Synthetic Vector Data Generation}
To generate high-dimensional vector data with a known causal structure, we first generate a random DAG and linear SCM parameters, and generate true causal variables by ancestral sampling. This is then used to generate corresponding high-dimensional dataset with a random projection function.

\textbf{Generating the DAG and causal variables}: Following many works in the literature, we sample random Erdős–Rényi (ER) DAGs \citep{erdos1960evolution} with degrees in $\{1, 2, 4\}$ to generate the DAG. For every edge in this DAG, we sample the magnitude of edge weights uniformly as $|L| \sim \mathcal{U}(0.5, 2.0)$ and randomly sample the permutation matrix. We perform ancestral sampling through this random DAG with intervention targets $\mathcal{I}$, to obtain $\textbf{Z}$ and then project it to $\mathrm{D}$ dimensions to obtain $\{ \mathbf{x}^{(i)}\}_{i=1}^N$. 

\textbf{Generating high-dimensional vectors from causal variables}: We consider two different cases of generating the high-dimensional data from the causal variables obtained in the previous step: (i) $\textbf{x}^{(i)}$ is a random linear projection of causal variables, $\textbf{z}^{(i)}$, from $\mathbb{R}^d$ to $\mathbb{R}^{\mathrm{D}}$, according to $\textbf{x}=\textbf{z}{\Tilde{\mathrm{P}}}$, where ${\Tilde{\mathrm{P}}} \in \mathbb{R}^{d \times \mathrm{D}}$ is a random projection matrix. (ii) $\textbf{x}^{(i)}$ is a nonlinear projection of causal variables, $\textbf{z}^{(i)}$, modeled by a 3-layer MLP. 

\begin{figure}[t] 
    \centering
    \includegraphics[width=14cm]{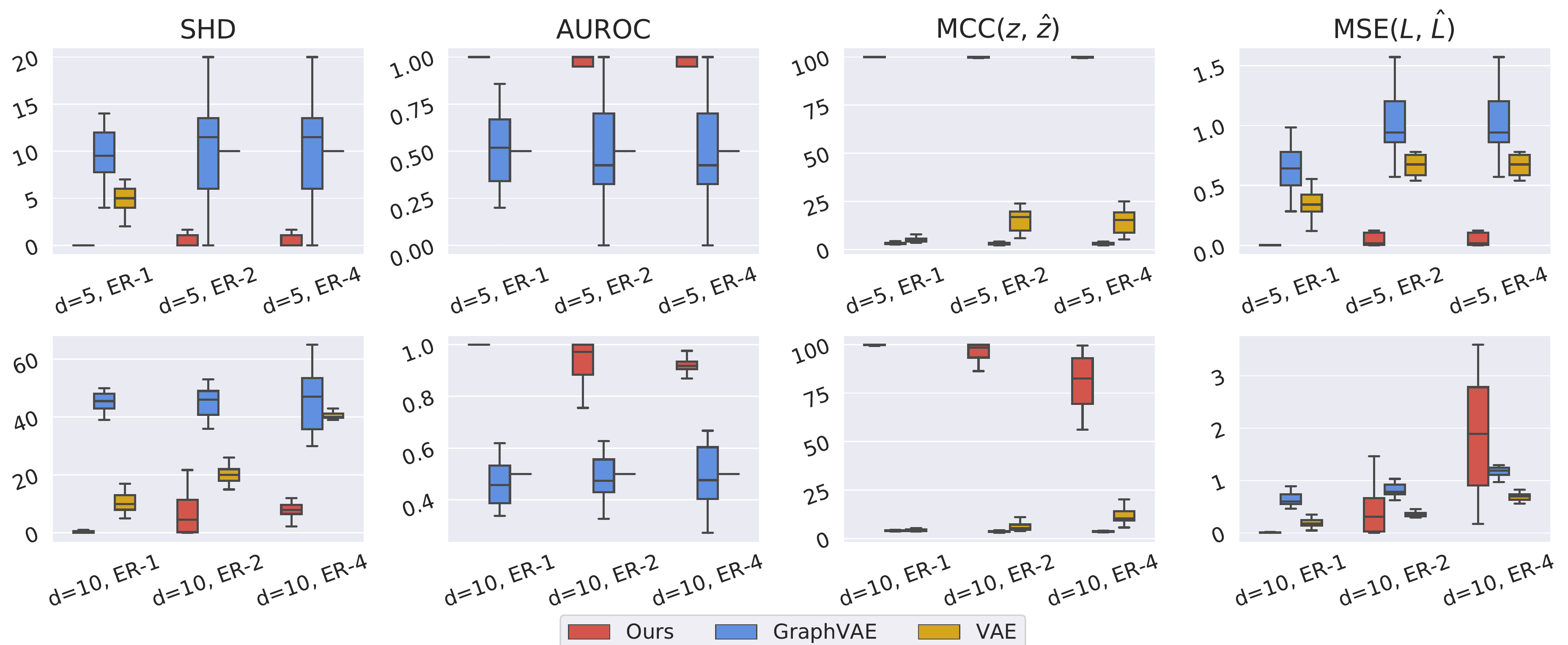}
    \vspace{-0.5cm}
    \caption{Learning the latent SCM for \textbf{nonlinear} projection of causal variables for $d=5$ (top) and $d=10$ (bottom) nodes, $\mathrm{D}=100$ dimensions. \textbf{$\mathbb{E}$-SHD ($\downarrow$)}, \textbf{AUROC ($\uparrow$)}, \textbf{MCC ($\uparrow$)}, \textbf{MSE ($\downarrow$)}.}
    \label{d5_d10_nonlinear_blcd}
\end{figure}

\subsubsection{Results on Synthetic Vector Data}
\textbf{Results on linear projection of causal variables}:  We present results on the learned causal variables, structure, and parameters in two scenarios: (i) when the true node ordering or permutation is given (e.g., as in \cite{graphvae}), and (ii) when the node ordering is \textit{not} given and one has to additionally also infer the permutation $P$. For $d=5, 10, 20$ nodes projected to $\mathrm{D}=100$ dimensions, we evaluate our algorithm on synthetic ER-1, ER-2, and ER-4 DAGs. The model was trained for 5000 epochs so as to reach convergence. Figure \ref{d5_lineardbcd} summarizes the results for $d=5$ nodes, for which we use 500 observational data points and 2000 interventional data points. Of the 2000 interventional data, we generate 100 random interventional data points per set over 20 intervention sets\footnote{An intervention set is defined as a set of nodes on which an intervention is performed}. It can be seen that when permutation is given, the proposed method can recover the causal graph correctly in all the cases, achieving $\E$-SHD of 0 and AUROC of 1. When the permutation is learned, the proposed method still recovers the true causal structure very well. However, this is not the case with baseline methods of VAE and GraphVAE, which perform significantly worse on most of the metrics. Figure \ref{d10_linear_dbcd_metrics} and \ref{d20_linear_dbcd_metrics} (in Appendix) show the results for $d=10$ and $d=20$ nodes.

\textbf{Results on nonlinear projection of causal variables}: For $d=5, 10, 20$ nodes projected to $\mathrm{D}=100$ dimensions, we evaluate our algorithm on synthetic ER-1, ER-2, and ER-4 DAGs, given the permutation $P$. Figure \ref{d5_d10_nonlinear_blcd} summarizes the results for 5 and 10 nodes. As in the linear case, the proposed method recovers the true causal structure and the true causal variables, and is significantly better than the VAE and GraphVAE baselines on all the metrics considered. For experiments in this setting, we noticed empirically that learning the permutation is hard, and performs not so different from a null graph baseline (Figure \ref{d6_nonlinear_learn_SCM}). This observation works that theoretically show the identifiability result that recovery of latent variables is possible only upto a permutation in latent causal models \citep{brehmer2022weakly, https://doi.org/10.48550/arxiv.2208.14153} for general nonlinear mappings between causal variables and low-level data. This supports our observation of not being able to learn the permutation in nonlinear projection settings -- but once the permutation is given to the model, it can quickly recover the SCM (figures \ref{d5_d10_nonlinear_blcd},\ref{d6_nonlinear_learn_SCM}). Refer figure \ref{d20_nonlinear} (in Appendix) for results on $d=20$ nodes. 

\subsection{Results on learning latent SCMs from pixel data} \label{setting_3}

\textbf{Dataset and Setup}: A major challenge with evaluating latent causal discovery models on images is that it is hard to obtain images with corresponding GT graph and parameters. Other works \citep{causalgan, causalvae, dear} handle this by assuming the dataset is generated from certain causal variables (assumed to be attributes like gender, baldness, etc.) and a causal structure that is heuristically set by experts, usually in the CelebA dataset \citep{celebA}. This makes evaluation particularly noisy. Given these limitations, we verify if our model can perform latent causal discovery by evaluating on images from the chemistry dataset proposed in \cite{ke2021systematic} -- a scenario where all GT factors are known. We use the environment to generate blocks of different intensities according to a linear Gaussian latent SCM where the parent block colors affect the child block colors then obtain the corresponding images of blocks. The dataset allows generating pixel data from random DAGs and linear SCMs. For this step, we use the same technique to generate causal variables as in the synthetic dataset section. 

\begin{figure}[t]
    \centering
    \includegraphics[width=14cm]{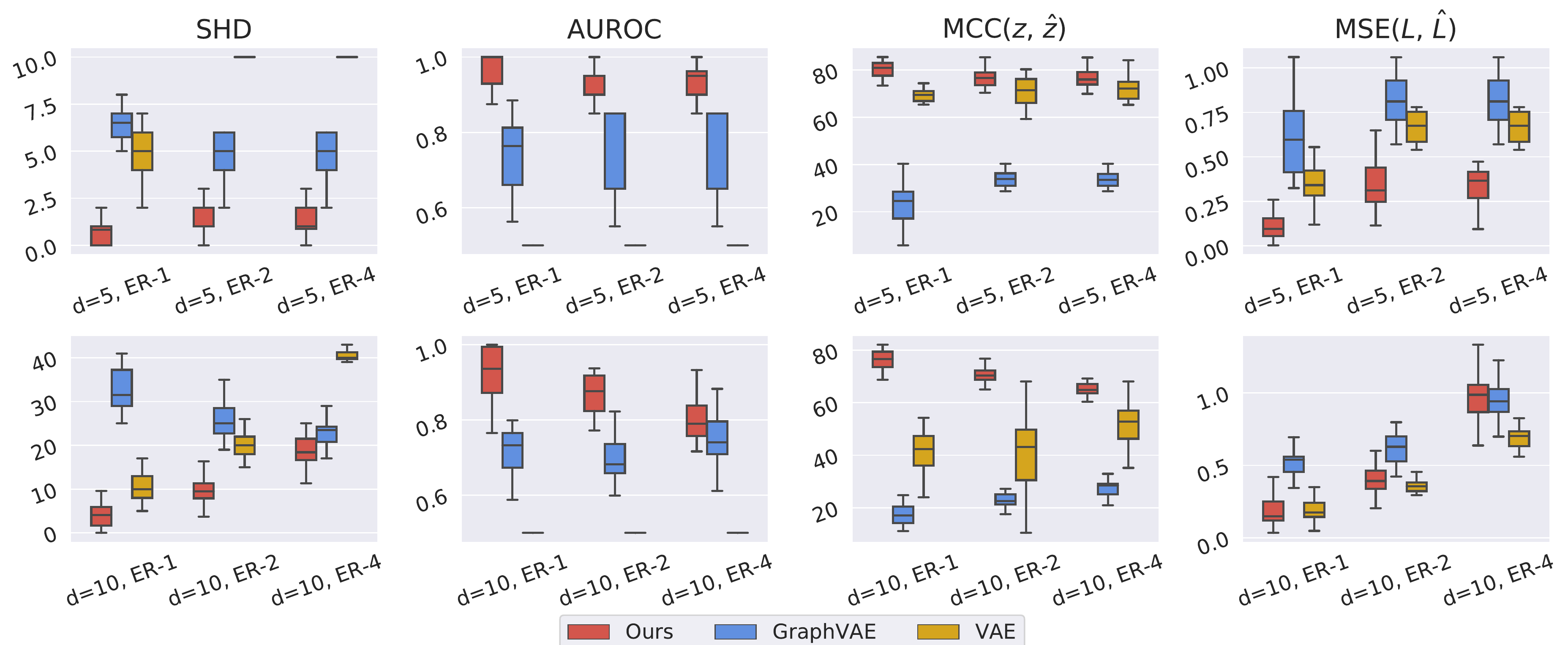}
    \vspace{-15pt}
    \caption{Learning the latent SCM from pixels of the chemistry dataset for $d = 5$ (top) and $d = 10$ nodes (bottom). \textbf{$\mathbb{E}$-SHD ($\downarrow$)}, \textbf{AUROC ($\uparrow$)}, \textbf{MCC ($\uparrow$)}, \textbf{MSE ($\downarrow$)}}
    \label{d5_10_chem_env_results}
\end{figure}

\begin{figure}[ht]
    \centering
    \includegraphics[width=14cm]{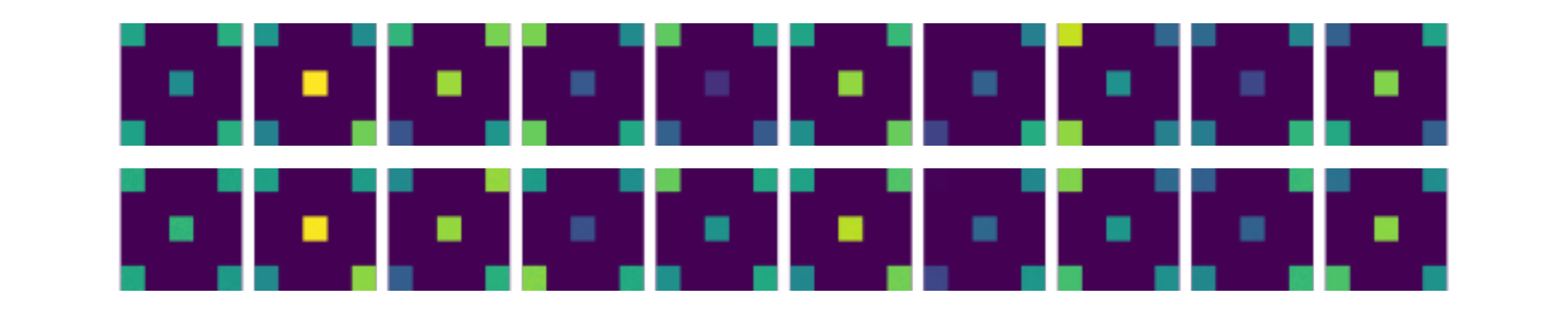}
    \vspace{-20pt}
    \caption{Image sampling from 10 random, unseen interventions: Mean over GT (top row) and predicted (bottom row) image samples in the chemistry dataset for  $d=5$ nodes.}
    \label{unseen_interv_d5}
\end{figure}

\textbf{Results}: We perform experiments to evaluate latent causal discovery from pixels and known interventions. The results are summarized in figure \ref{d5_10_chem_env_results}. It can be seen that the proposed approach can recover the SCM significantly better than the baseline approaches in all the metrics even in the realistic dataset. In figure \ref{unseen_interv_d5}, we also assess the ability of the model to sample images from unseen interventions in the chemistry dataset by examining the generated images with GT interventional samples. The matching intensity of each block corresponds to matching causal variables, which demonstrates model generalization.

\section{Conclusion}
We presented a tractable approximate inference technique to perform Bayesian latent causal discovery that jointly infers the causal variables, structure and parameters of linear Gaussian latent SCMs under random, known interventions from low-level data. The learned causal model is also shown to generalize to unseen interventions. Our Bayesian formulation allows uncertainty quantification and mutual information estimation which is well-suited for extensions to active causal discovery. Extensions of the proposed method to learn nonlinear, non-Gaussian latent SCMs from unknown interventions would also open doors to general algorithms that can learn causal representations.

\section{Acknowledgements and disclosure of funding}
The authors thank Anirudh Goyal and Sébastien Lachapelle for fruitful discussions and feedback on this work, and to Nanda Harishankar Krishna for reviewing early versions of this paper. We are also thankful to Compute Canada and CIFAR for the compute and funding that made this work possible.


\bibliography{iclr2023_conference}

\begin{thebibliography}{77}
\providecommand{\natexlab}[1]{#1}
\providecommand{\url}[1]{\texttt{#1}}
\expandafter\ifx\csname urlstyle\endcsname\relax
  \providecommand{\doi}[1]{doi: #1}\else
  \providecommand{\doi}{doi: \begingroup \urlstyle{rm}\Url}\fi

\bibitem[Agrawal et~al.(2019)Agrawal, Squires, Yang, Shanmugam, and
  Uhler]{abcd}
Raj Agrawal, Chandler Squires, Karren Yang, Karthikeyan Shanmugam, and Caroline
  Uhler.
\newblock Abcd-strategy: Budgeted experimental design for targeted causal
  structure discovery.
\newblock In Kamalika Chaudhuri and Masashi Sugiyama, editors,
  \emph{Proceedings of the Twenty-Second International Conference on Artificial
  Intelligence and Statistics}, volume~89 of \emph{Proceedings of Machine
  Learning Research}, pages 3400--3409. PMLR, 16--18 Apr 2019.
\newblock URL \url{https://proceedings.mlr.press/v89/agrawal19b.html}.

\bibitem[Ahuja et~al.(2022)Ahuja, Mahajan, Syrgkanis, and
  Mitliagkas]{https://doi.org/10.48550/arxiv.2204.04606}
Kartik Ahuja, Divyat Mahajan, Vasilis Syrgkanis, and Ioannis Mitliagkas.
\newblock Towards efficient representation identification in supervised
  learning, 2022.
\newblock URL \url{https://arxiv.org/abs/2204.04606}.

\bibitem[Annadani et~al.(2021)Annadani, Rothfuss, Lacoste, Scherrer, Goyal,
  Bengio, and Bauer]{vcn}
Yashas Annadani, Jonas Rothfuss, Alexandre Lacoste, Nino Scherrer, Anirudh
  Goyal, Yoshua Bengio, and Stefan Bauer.
\newblock Variational causal networks: Approximate bayesian inference over
  causal structures.
\newblock \emph{arXiv preprint arXiv:2106.07635}, 2021.

\bibitem[Barron et~al.(1998)Barron, Rissanen, and Yu]{720554}
A.~Barron, J.~Rissanen, and Bin Yu.
\newblock The minimum description length principle in coding and modeling.
\newblock \emph{IEEE Transactions on Information Theory}, 44\penalty0
  (6):\penalty0 2743--2760, 1998.
\newblock \doi{10.1109/18.720554}.

\bibitem[Bengio et~al.(2021)Bengio, Jain, Korablyov, Precup, and
  Bengio]{bengio2021flow}
Emmanuel Bengio, Moksh Jain, Maksym Korablyov, Doina Precup, and Yoshua Bengio.
\newblock Flow network based generative models for non-iterative diverse
  candidate generation.
\newblock \emph{Advances in Neural Information Processing Systems},
  34:\penalty0 27381--27394, 2021.

\bibitem[Bengio et~al.(2012)Bengio, Courville, and Vincent]{replearn}
Yoshua Bengio, Aaron Courville, and Pascal Vincent.
\newblock Representation learning: A review and new perspectives, 2012.
\newblock URL \url{https://arxiv.org/abs/1206.5538}.

\bibitem[Bengio et~al.(2019)Bengio, Deleu, Rahaman, Ke, Lachapelle, Bilaniuk,
  Goyal, and Pal]{bengio2019meta}
Yoshua Bengio, Tristan Deleu, Nasim Rahaman, Rosemary Ke, Sébastien
  Lachapelle, Olexa Bilaniuk, Anirudh Goyal, and Christopher Pal.
\newblock A meta-transfer objective for learning to disentangle causal
  mechanisms, 2019.
\newblock URL \url{https://arxiv.org/abs/1901.10912}.

\bibitem[Blei et~al.(2017)Blei, Kucukelbir, and McAuliffe]{blei2017variational}
David~M Blei, Alp Kucukelbir, and Jon~D McAuliffe.
\newblock Variational inference: A review for statisticians.
\newblock \emph{Journal of the American statistical Association}, 112\penalty0
  (518):\penalty0 859--877, 2017.

\bibitem[Bradbury et~al.(2018)Bradbury, Frostig, Hawkins, Johnson, Leary,
  Maclaurin, Necula, Paszke, VanderPlas, Wanderman-Milne,
  et~al.]{jax2018github}
James Bradbury, Roy Frostig, Peter Hawkins, Matthew~James Johnson, Chris Leary,
  Dougal Maclaurin, George Necula, Adam Paszke, Jake VanderPlas, Skye
  Wanderman-Milne, et~al.
\newblock Jax: composable transformations of python+ numpy programs.
\newblock \emph{Version 0.2}, 5:\penalty0 14--24, 2018.

\bibitem[Brehmer et~al.(2022)Brehmer, De~Haan, Lippe, and
  Cohen]{brehmer2022weakly}
Johann Brehmer, Pim De~Haan, Phillip Lippe, and Taco Cohen.
\newblock Weakly supervised causal representation learning.
\newblock \emph{arXiv preprint arXiv:2203.16437}, 2022.

\bibitem[Brouillard et~al.(2020)Brouillard, Lachapelle, Lacoste,
  Lacoste-Julien, and Drouin]{DCDI}
Philippe Brouillard, S{\'e}bastien Lachapelle, Alexandre Lacoste, Simon
  Lacoste-Julien, and Alexandre Drouin.
\newblock Differentiable causal discovery from interventional data.
\newblock \emph{Advances in Neural Information Processing Systems},
  33:\penalty0 21865--21877, 2020.

\bibitem[Carvalho et~al.(2009)Carvalho, Polson, and Scott]{horseshoe}
Carlos~M. Carvalho, Nicholas~G. Polson, and James~G. Scott.
\newblock Handling sparsity via the horseshoe.
\newblock In David van Dyk and Max Welling, editors, \emph{Proceedings of the
  Twelth International Conference on Artificial Intelligence and Statistics},
  volume~5 of \emph{Proceedings of Machine Learning Research}, pages 73--80,
  Hilton Clearwater Beach Resort, Clearwater Beach, Florida USA, 16--18 Apr
  2009. PMLR.
\newblock URL \url{https://proceedings.mlr.press/v5/carvalho09a.html}.

\bibitem[Charpentier et~al.(2022)Charpentier, Kibler, and
  G{\"u}nnemann]{diff_dag_sampling}
Bertrand Charpentier, Simon Kibler, and Stephan G{\"u}nnemann.
\newblock Differentiable dag sampling.
\newblock \emph{arXiv preprint arXiv:2203.08509}, 2022.

\bibitem[Cheng et~al.(2002)Cheng, Greiner, Kelly, Bell, and Liu]{CHENG200243}
Jie Cheng, Russell Greiner, Jonathan Kelly, David Bell, and Weiru Liu.
\newblock Learning bayesian networks from data: An information-theory based
  approach.
\newblock \emph{Artificial Intelligence}, 137\penalty0 (1):\penalty0 43--90,
  2002.
\newblock ISSN 0004-3702.
\newblock \doi{https://doi.org/10.1016/S0004-3702(02)00191-1}.
\newblock URL
  \url{https://www.sciencedirect.com/science/article/pii/S0004370202001911}.

\bibitem[Chickering(2002)]{ges}
David~Maxwell Chickering.
\newblock Optimal structure identification with greedy search.
\newblock \emph{Journal of machine learning research}, 3\penalty0
  (Nov):\penalty0 507--554, 2002.

\bibitem[Cundy et~al.(2021)Cundy, Grover, and Ermon]{bcdnets}
Chris Cundy, Aditya Grover, and Stefano Ermon.
\newblock Bcd nets: Scalable variational approaches for bayesian causal
  discovery.
\newblock In M.~Ranzato, A.~Beygelzimer, Y.~Dauphin, P.S. Liang, and J.~Wortman
  Vaughan, editors, \emph{Advances in Neural Information Processing Systems},
  volume~34, pages 7095--7110. Curran Associates, Inc., 2021.
\newblock URL
  \url{https://proceedings.neurips.cc/paper/2021/file/39799c18791e8d7eb29704fc5bc04ac8-Paper.pdf}.

\bibitem[Deleu et~al.(2022)Deleu, G{\'o}is, Emezue, Rankawat, Lacoste-Julien,
  Bauer, and Bengio]{daggfn}
Tristan Deleu, Ant{\'o}nio G{\'o}is, Chris Emezue, Mansi Rankawat, Simon
  Lacoste-Julien, Stefan Bauer, and Yoshua Bengio.
\newblock Bayesian structure learning with generative flow networks.
\newblock \emph{arXiv preprint arXiv:2202.13903}, 2022.

\bibitem[Erdos et~al.(1960)Erdos, R{\'e}nyi, et~al.]{erdos1960evolution}
Paul Erdos, Alfr{\'e}d R{\'e}nyi, et~al.
\newblock On the evolution of random graphs.
\newblock \emph{Publ. Math. Inst. Hung. Acad. Sci}, 5\penalty0 (1):\penalty0
  17--60, 1960.

\bibitem[Friedman and Koller(2013)]{friedman2013being}
Nir Friedman and Daphne Koller.
\newblock Being bayesian about network structure.
\newblock \emph{arXiv preprint arXiv:1301.3856}, 2013.

\bibitem[Geiger and Heckerman(1994)]{Geiger1994LearningGN}
Dan Geiger and David Heckerman.
\newblock Learning gaussian networks.
\newblock In \emph{UAI}, 1994.

\bibitem[Ghoshal and Honorio(2018)]{ghoshal2018learning}
Asish Ghoshal and Jean Honorio.
\newblock Learning linear structural equation models in polynomial time and
  sample complexity.
\newblock In \emph{International Conference on Artificial Intelligence and
  Statistics}, pages 1466--1475. PMLR, 2018.

\bibitem[H{\"a}gele et~al.(2022)H{\"a}gele, Rothfuss, Lorch, Somnath,
  Sch{\"o}lkopf, and Krause]{hagele2022bacadi}
Alexander H{\"a}gele, Jonas Rothfuss, Lars Lorch, Vignesh~Ram Somnath, Bernhard
  Sch{\"o}lkopf, and Andreas Krause.
\newblock Bacadi: Bayesian causal discovery with unknown interventions.
\newblock \emph{arXiv preprint arXiv:2206.01665}, 2022.

\bibitem[He et~al.(2019)He, Gong, Marino, Mori, and Lehrmann]{graphvae}
Jiawei He, Yu~Gong, Joseph Marino, Greg Mori, and Andreas Lehrmann.
\newblock Variational autoencoders with jointly optimized latent dependency
  structure.
\newblock In \emph{International Conference on Learning Representations}, 2019.
\newblock URL \url{https://openreview.net/forum?id=SJgsCjCqt7}.

\bibitem[Heckerman et~al.(1995)Heckerman, Geiger, and
  Chickering]{heckerman1995learning}
David Heckerman, Dan Geiger, and David~M Chickering.
\newblock Learning bayesian networks: The combination of knowledge and
  statistical data.
\newblock \emph{Machine learning}, 20\penalty0 (3):\penalty0 197--243, 1995.

\bibitem[Heckerman et~al.(1997)Heckerman, Meek, and
  Cooper]{heckerman1997bayesian}
David Heckerman, Christopher Meek, and Gregory Cooper.
\newblock A bayesian approach to causal discovery.
\newblock Technical report, Technical report msr-tr-97-05, Microsoft Research,
  1997.

\bibitem[Heckerman et~al.(2006)Heckerman, Meek, and Cooper]{bacd}
David Heckerman, Christopher Meek, and Gregory Cooper.
\newblock A bayesian approach to causal discovery.
\newblock In \emph{Innovations in Machine Learning}, pages 1--28. Springer,
  2006.

\bibitem[Hyvarinen and Morioka(2017)]{pmlr-v54-hyvarinen17a}
Aapo Hyvarinen and Hiroshi Morioka.
\newblock {Nonlinear ICA of Temporally Dependent Stationary Sources}.
\newblock In Aarti Singh and Jerry Zhu, editors, \emph{Proceedings of the 20th
  International Conference on Artificial Intelligence and Statistics},
  volume~54 of \emph{Proceedings of Machine Learning Research}, pages 460--469.
  PMLR, 20--22 Apr 2017.
\newblock URL \url{https://proceedings.mlr.press/v54/hyvarinen17a.html}.

\bibitem[Kass and Raftery(1995)]{10.2307/2291091}
Robert~E. Kass and Adrian~E. Raftery.
\newblock Bayes factors.
\newblock \emph{Journal of the American Statistical Association}, 90\penalty0
  (430):\penalty0 773--795, 1995.
\newblock ISSN 01621459.
\newblock URL \url{http://www.jstor.org/stable/2291091}.

\bibitem[Ke et~al.(2019)Ke, Bilaniuk, Goyal, Bauer, Larochelle, Sch{\"o}lkopf,
  Mozer, Pal, and Bengio]{ke2019learning}
Nan~Rosemary Ke, Olexa Bilaniuk, Anirudh Goyal, Stefan Bauer, Hugo Larochelle,
  Bernhard Sch{\"o}lkopf, Michael~C Mozer, Chris Pal, and Yoshua Bengio.
\newblock Learning neural causal models from unknown interventions.
\newblock \emph{arXiv preprint arXiv:1910.01075}, 2019.

\bibitem[Ke et~al.(2021)Ke, Didolkar, Mittal, Goyal, Lajoie, Bauer, Rezende,
  Bengio, Mozer, and Pal]{ke2021systematic}
Nan~Rosemary Ke, Aniket Didolkar, Sarthak Mittal, Anirudh Goyal, Guillaume
  Lajoie, Stefan Bauer, Danilo Rezende, Yoshua Bengio, Michael Mozer, and
  Christopher Pal.
\newblock Systematic evaluation of causal discovery in visual model based
  reinforcement learning.
\newblock \emph{arXiv preprint arXiv:2107.00848}, 2021.

\bibitem[Ke et~al.(2022)Ke, Chiappa, Wang, Bornschein, Weber, Goyal, Botvinic,
  Mozer, and Rezende]{ke2022learning}
Nan~Rosemary Ke, Silvia Chiappa, Jane Wang, Jorg Bornschein, Theophane Weber,
  Anirudh Goyal, Matthew Botvinic, Michael Mozer, and Danilo~Jimenez Rezende.
\newblock Learning to induce causal structure.
\newblock \emph{arXiv preprint arXiv:2204.04875}, 2022.

\bibitem[Kingma and Welling(2013)]{vae}
Diederik~P Kingma and Max Welling.
\newblock Auto-encoding variational bayes, 2013.
\newblock URL \url{https://arxiv.org/abs/1312.6114}.

\bibitem[Kingma et~al.(2016)Kingma, Salimans, Jozefowicz, Chen, Sutskever, and
  Welling]{vae_iaf}
Diederik~P. Kingma, Tim Salimans, Rafal Jozefowicz, Xi~Chen, Ilya Sutskever,
  and Max Welling.
\newblock Improving variational inference with inverse autoregressive flow,
  2016.
\newblock URL \url{https://arxiv.org/abs/1606.04934}.

\bibitem[Kocaoglu et~al.(2017)Kocaoglu, Snyder, Dimakis, and
  Vishwanath]{causalgan}
Murat Kocaoglu, Christopher Snyder, Alexandros~G. Dimakis, and Sriram
  Vishwanath.
\newblock Causalgan: Learning causal implicit generative models with
  adversarial training, 2017.
\newblock URL \url{https://arxiv.org/abs/1709.02023}.

\bibitem[Koivisto and Sood(2004)]{koivisto2004exact}
Mikko Koivisto and Kismat Sood.
\newblock Exact bayesian structure discovery in bayesian networks.
\newblock \emph{The Journal of Machine Learning Research}, 5:\penalty0
  549--573, 2004.

\bibitem[Kuhn(1955)]{hungarian}
H.~W. Kuhn.
\newblock The hungarian method for the assignment problem.
\newblock \emph{Naval Research Logistics Quarterly}, 2\penalty0 (1-2):\penalty0
  83--97, 1955.
\newblock \doi{https://doi.org/10.1002/nav.3800020109}.
\newblock URL
  \url{https://onlinelibrary.wiley.com/doi/abs/10.1002/nav.3800020109}.

\bibitem[Kuipers et~al.(2022)Kuipers, Suter, and Moffa]{kuipers2022efficient}
Jack Kuipers, Polina Suter, and Giusi Moffa.
\newblock Efficient sampling and structure learning of bayesian networks.
\newblock \emph{Journal of Computational and Graphical Statistics}, pages
  1--12, 2022.

\bibitem[Lachapelle et~al.(2019)Lachapelle, Brouillard, Deleu, and
  Lacoste-Julien]{grandag}
S{\'e}bastien Lachapelle, Philippe Brouillard, Tristan Deleu, and Simon
  Lacoste-Julien.
\newblock Gradient-based neural dag learning.
\newblock \emph{arXiv preprint arXiv:1906.02226}, 2019.

\bibitem[Lehmann and Romano(2005)]{lehmann2005testing}
E.~L. Lehmann and Joseph~P. Romano.
\newblock \emph{Testing statistical hypotheses}.
\newblock Springer Texts in Statistics. Springer, New York, third edition,
  2005.
\newblock ISBN 0-387-98864-5.

\bibitem[Li et~al.(2022)Li, Li, Wang, Wei, Pfister, and Chan]{ircm}
Yuke Li, Kenneth Li, Pin Wang, Donglai Wei, Hanspeter Pfister, and Ching-Yao
  Chan.
\newblock Intervention-based recurrent causal model for non-stationary video
  causal discovery, 2022.
\newblock URL \url{https://openreview.net/forum?id=JvGzKO1QLet}.

\bibitem[Lippe et~al.(2022)Lippe, Magliacane, L{\"o}we, Asano, Cohen, and
  Gavves]{citris}
Phillip Lippe, Sara Magliacane, Sindy L{\"o}we, Yuki~M Asano, Taco Cohen, and
  Stratis Gavves.
\newblock Citris: Causal identifiability from temporal intervened sequences.
\newblock In \emph{International Conference on Machine Learning}, pages
  13557--13603. PMLR, 2022.

\bibitem[Liu et~al.(2022)Liu, Zhang, Gong, Gong, Huang, Hengel, Zhang, and
  Shi]{https://doi.org/10.48550/arxiv.2208.14153}
Yuhang Liu, Zhen Zhang, Dong Gong, Mingming Gong, Biwei Huang, Anton van~den
  Hengel, Kun Zhang, and Javen~Qinfeng Shi.
\newblock Weight-variant latent causal models, 2022.
\newblock URL \url{https://arxiv.org/abs/2208.14153}.

\bibitem[Liu et~al.(2015)Liu, Luo, Wang, and Tang]{celebA}
Ziwei Liu, Ping Luo, Xiaogang Wang, and Xiaoou Tang.
\newblock Deep learning face attributes in the wild.
\newblock In \emph{Proceedings of the IEEE international conference on computer
  vision}, pages 3730--3738, 2015.

\bibitem[Lopez-Paz and Oquab(2016)]{rctt}
David Lopez-Paz and Maxime Oquab.
\newblock Revisiting classifier two-sample tests, 2016.
\newblock URL \url{https://arxiv.org/abs/1610.06545}.

\bibitem[Lopez-Paz et~al.(2016)Lopez-Paz, Nishihara, Chintala, Schölkopf, and
  Bottou]{csii}
David Lopez-Paz, Robert Nishihara, Soumith Chintala, Bernhard Schölkopf, and
  Léon Bottou.
\newblock Discovering causal signals in images, 2016.
\newblock URL \url{https://arxiv.org/abs/1605.08179}.

\bibitem[Lorch et~al.(2021)Lorch, Rothfuss, Sch\"{o}lkopf, and Krause]{dibs}
Lars Lorch, Jonas Rothfuss, Bernhard Sch\"{o}lkopf, and Andreas Krause.
\newblock Dibs: Differentiable bayesian structure learning.
\newblock In M.~Ranzato, A.~Beygelzimer, Y.~Dauphin, P.S. Liang, and J.~Wortman
  Vaughan, editors, \emph{Advances in Neural Information Processing Systems},
  volume~34, pages 24111--24123. Curran Associates, Inc., 2021.
\newblock URL
  \url{https://proceedings.neurips.cc/paper/2021/file/ca6ab34959489659f8c3776aaf1f8efd-Paper.pdf}.

\bibitem[Mena et~al.(2018)Mena, Belanger, Linderman, and
  Snoek]{gs_perm_learning}
Gonzalo Mena, David Belanger, Scott Linderman, and Jasper Snoek.
\newblock Learning latent permutations with gumbel-sinkhorn networks.
\newblock \emph{arXiv preprint arXiv:1802.08665}, 2018.

\bibitem[Moraffah et~al.(2020)Moraffah, Moraffah, Karami, Raglin, and Liu]{can}
Raha Moraffah, Bahman Moraffah, Mansooreh Karami, Adrienne Raglin, and Huan
  Liu.
\newblock Causal adversarial network for learning conditional and
  interventional distributions, 2020.
\newblock URL \url{https://arxiv.org/abs/2008.11376}.

\bibitem[Ng et~al.(2020)Ng, Ghassami, and Zhang]{ng2020role}
Ignavier Ng, AmirEmad Ghassami, and Kun Zhang.
\newblock On the role of sparsity and dag constraints for learning linear dags.
\newblock \emph{Advances in Neural Information Processing Systems},
  33:\penalty0 17943--17954, 2020.

\bibitem[Ng et~al.(2022)Ng, Zhu, Fang, Li, Chen, and Wang]{ng2022masked}
Ignavier Ng, Shengyu Zhu, Zhuangyan Fang, Haoyang Li, Zhitang Chen, and Jun
  Wang.
\newblock Masked gradient-based causal structure learning.
\newblock In \emph{Proceedings of the 2022 SIAM International Conference on
  Data Mining (SDM)}, pages 424--432. SIAM, 2022.

\bibitem[Pamfil et~al.(2020)Pamfil, Sriwattanaworachai, Desai, Pilgerstorfer,
  Georgatzis, Beaumont, and Aragam]{dynotears}
Roxana Pamfil, Nisara Sriwattanaworachai, Shaan Desai, Philip Pilgerstorfer,
  Konstantinos Georgatzis, Paul Beaumont, and Bryon Aragam.
\newblock Dynotears: Structure learning from time-series data.
\newblock In \emph{International Conference on Artificial Intelligence and
  Statistics}, pages 1595--1605. PMLR, 2020.

\bibitem[Pearl(2009)]{pearl_2009}
Judea Pearl.
\newblock \emph{Causality}.
\newblock Cambridge University Press, 2 edition, 2009.
\newblock \doi{10.1017/CBO9780511803161}.

\bibitem[Peters and B{\"u}hlmann(2014)]{gdseev}
Jonas Peters and Peter B{\"u}hlmann.
\newblock Identifiability of gaussian structural equation models with equal
  error variances.
\newblock \emph{Biometrika}, 101\penalty0 (1):\penalty0 219--228, 2014.

\bibitem[Rezende et~al.(2014)Rezende, Mohamed, and Wierstra]{vae_rezende}
Danilo~Jimenez Rezende, Shakir Mohamed, and Daan Wierstra.
\newblock Stochastic backpropagation and approximate inference in deep
  generative models, 2014.
\newblock URL \url{https://arxiv.org/abs/1401.4082}.

\bibitem[Scherrer et~al.(2021)Scherrer, Bilaniuk, Annadani, Goyal, Schwab,
  Sch{\"o}lkopf, Mozer, Bengio, Bauer, and Ke]{scherrer2021learning}
Nino Scherrer, Olexa Bilaniuk, Yashas Annadani, Anirudh Goyal, Patrick Schwab,
  Bernhard Sch{\"o}lkopf, Michael~C Mozer, Yoshua Bengio, Stefan Bauer, and
  Nan~Rosemary Ke.
\newblock Learning neural causal models with active interventions.
\newblock \emph{arXiv preprint arXiv:2109.02429}, 2021.

\bibitem[Scherrer et~al.(2022)Scherrer, Goyal, Bauer, Bengio, and Ke]{nino_ood}
Nino Scherrer, Anirudh Goyal, Stefan Bauer, Yoshua Bengio, and Nan~Rosemary Ke.
\newblock On the generalization and adaption performance of causal models,
  2022.
\newblock URL \url{https://arxiv.org/abs/2206.04620}.

\bibitem[Schoelkopf et~al.(2012)Schoelkopf, Janzing, Peters, Sgouritsa, Zhang,
  and Mooij]{ocacl}
Bernhard Schoelkopf, Dominik Janzing, Jonas Peters, Eleni Sgouritsa, Kun Zhang,
  and Joris Mooij.
\newblock On causal and anticausal learning, 2012.
\newblock URL \url{https://arxiv.org/abs/1206.6471}.

\bibitem[Schölkopf et~al.(2021)Schölkopf, Locatello, Bauer, Ke, Kalchbrenner,
  Goyal, and Bengio]{tcrl}
Bernhard Schölkopf, Francesco Locatello, Stefan Bauer, Nan~Rosemary Ke, Nal
  Kalchbrenner, Anirudh Goyal, and Yoshua Bengio.
\newblock Towards causal representation learning, 2021.
\newblock URL \url{https://arxiv.org/abs/2102.11107}.

\bibitem[Shen et~al.(2020)Shen, Liu, Dong, Lian, Chen, and Zhang]{dear}
Xinwei Shen, Furui Liu, Hanze Dong, Qing Lian, Zhitang Chen, and Tong Zhang.
\newblock Disentangled generative causal representation learning, 2020.
\newblock URL \url{https://arxiv.org/abs/2010.02637}.

\bibitem[Shimizu et~al.(2011)Shimizu, Inazumi, Sogawa, Hyv{\"a}rinen, Kawahara,
  Washio, Hoyer, and Bollen]{shimizu2011directlingam}
Shohei Shimizu, Takanori Inazumi, Yasuhiro Sogawa, Aapo Hyv{\"a}rinen,
  Yoshinobu Kawahara, Takashi Washio, Patrik~O Hoyer, and Kenneth Bollen.
\newblock Directlingam: A direct method for learning a linear non-gaussian
  structural equation model.
\newblock \emph{The Journal of Machine Learning Research}, 12:\penalty0
  1225--1248, 2011.

\bibitem[Spirtes et~al.(2000)Spirtes, Glymour, and Scheines]{pc}
P.~Spirtes, C.~Glymour, and R.~Scheines.
\newblock \emph{Causation, Prediction, and Search}.
\newblock MIT press, 2nd edition, 2000.

\bibitem[Sønderby et~al.(2016)Sønderby, Raiko, Maaløe, Sønderby, and
  Winther]{laddervae}
Casper~Kaae Sønderby, Tapani Raiko, Lars Maaløe, Søren~Kaae Sønderby, and
  Ole Winther.
\newblock Ladder variational autoencoders, 2016.
\newblock URL \url{https://arxiv.org/abs/1602.02282}.

\bibitem[Tigas et~al.(2022)Tigas, Annadani, Jesson, Sch{\"o}lkopf, Gal, and
  Bauer]{tigas2022interventions}
Panagiotis Tigas, Yashas Annadani, Andrew Jesson, Bernhard Sch{\"o}lkopf, Yarin
  Gal, and Stefan Bauer.
\newblock Interventions, where and how? experimental design for causal models
  at scale.
\newblock \emph{arXiv preprint arXiv:2203.02016}, 2022.

\bibitem[Toth et~al.(2022)Toth, Lorch, Knoll, Krause, Pernkopf, Peharz, and von
  K{\"u}gelgen]{toth2022active}
Christian Toth, Lars Lorch, Christian Knoll, Andreas Krause, Franz Pernkopf,
  Robert Peharz, and Julius von K{\"u}gelgen.
\newblock Active bayesian causal inference.
\newblock \emph{arXiv preprint arXiv:2206.02063}, 2022.

\bibitem[Vaswani et~al.(2017)Vaswani, Shazeer, Parmar, Uszkoreit, Jones, Gomez,
  Kaiser, and Polosukhin]{vaswani2017attention}
Ashish Vaswani, Noam Shazeer, Niki Parmar, Jakob Uszkoreit, Llion Jones,
  Aidan~N Gomez, {\L}ukasz Kaiser, and Illia Polosukhin.
\newblock Attention is all you need.
\newblock \emph{Advances in neural information processing systems}, 30, 2017.

\bibitem[Viinikka et~al.(2020)Viinikka, Hyttinen, Pensar, and Koivisto]{gadget}
Jussi Viinikka, Antti Hyttinen, Johan Pensar, and Mikko Koivisto.
\newblock Towards scalable bayesian learning of causal dags.
\newblock \emph{Advances in Neural Information Processing Systems},
  33:\penalty0 6584--6594, 2020.

\bibitem[Wang et~al.(2022)Wang, Wicker, and Kwiatkowska]{trust}
Benjie Wang, Matthew~R Wicker, and Marta Kwiatkowska.
\newblock Tractable uncertainty for structure learning.
\newblock In \emph{International Conference on Machine Learning}, pages
  23131--23150. PMLR, 2022.

\bibitem[Wang and Jordan(2021)]{causal_desiderata}
Yixin Wang and Michael~I. Jordan.
\newblock Desiderata for representation learning: A causal perspective, 2021.
\newblock URL \url{https://arxiv.org/abs/2109.03795}.

\bibitem[Wei et~al.(2020)Wei, Gao, and Yu]{nofears}
Dennis Wei, Tian Gao, and Yue Yu.
\newblock Dags with no fears: A closer look at continuous optimization for
  learning bayesian networks.
\newblock \emph{Advances in Neural Information Processing Systems},
  33:\penalty0 3895--3906, 2020.

\bibitem[Xie et~al.(2022)Xie, Huang, Chen, He, Geng, and Zhang]{LaHME}
Feng Xie, Biwei Huang, Zhengming Chen, Yangbo He, Zhi Geng, and Kun Zhang.
\newblock Identification of linear non-{G}aussian latent hierarchical
  structure.
\newblock In Kamalika Chaudhuri, Stefanie Jegelka, Le~Song, Csaba Szepesvari,
  Gang Niu, and Sivan Sabato, editors, \emph{Proceedings of the 39th
  International Conference on Machine Learning}, volume 162 of
  \emph{Proceedings of Machine Learning Research}, pages 24370--24387. PMLR,
  17--23 Jul 2022.
\newblock URL \url{https://proceedings.mlr.press/v162/xie22a.html}.

\bibitem[Yang et~al.(2021)Yang, Liu, Chen, Shen, Hao, and Wang]{causalvae}
Mengyue Yang, Furui Liu, Zhitang Chen, Xinwei Shen, Jianye Hao, and Jun Wang.
\newblock Causalvae: Disentangled representation learning via neural structural
  causal models.
\newblock In \emph{Proceedings of the IEEE/CVF Conference on Computer Vision
  and Pattern Recognition}, pages 9593--9602, 2021.

\bibitem[Yu et~al.(2019)Yu, Chen, Gao, and Yu]{dag_gnn}
Yue Yu, Jie Chen, Tian Gao, and Mo~Yu.
\newblock Dag-gnn: Dag structure learning with graph neural networks.
\newblock In \emph{International Conference on Machine Learning}, pages
  7154--7163. PMLR, 2019.

\bibitem[Yu et~al.(2021)Yu, Gao, Yin, and Ji]{dagnocurl}
Yue Yu, Tian Gao, Naiyu Yin, and Qiang Ji.
\newblock Dags with no curl: An efficient dag structure learning approach,
  2021.
\newblock URL \url{https://arxiv.org/abs/2106.07197}.

\bibitem[Zhang et~al.(2022)Zhang, Ng, Gong, Liu, Abbasnejad, Gong, Zhang, and
  Shi]{https://doi.org/10.48550/arxiv.2208.14571}
Zhen Zhang, Ignavier Ng, Dong Gong, Yuhang Liu, Ehsan~M Abbasnejad, Mingming
  Gong, Kun Zhang, and Javen~Qinfeng Shi.
\newblock Truncated matrix power iteration for differentiable dag learning,
  2022.
\newblock URL \url{https://arxiv.org/abs/2208.14571}.

\bibitem[Zhao et~al.(2017)Zhao, Song, and Ermon]{zhao}
Shengjia Zhao, Jiaming Song, and Stefano Ermon.
\newblock Learning hierarchical features from deep generative models.
\newblock In Doina Precup and Yee~Whye Teh, editors, \emph{Proceedings of the
  34th International Conference on Machine Learning}, volume~70 of
  \emph{Proceedings of Machine Learning Research}, pages 4091--4099. PMLR,
  06--11 Aug 2017.
\newblock URL \url{https://proceedings.mlr.press/v70/zhao17c.html}.

\bibitem[Zheng et~al.(2018)Zheng, Aragam, Ravikumar, and Xing]{zheng2018dags}
Xun Zheng, Bryon Aragam, Pradeep~K Ravikumar, and Eric~P Xing.
\newblock Dags with no tears: Continuous optimization for structure learning.
\newblock \emph{Advances in Neural Information Processing Systems}, 31, 2018.

\bibitem[Zimmermann et~al.(2021)Zimmermann, Sharma, Schneider, Bethge, and
  Brendel]{mcc2}
Roland~S. Zimmermann, Yash Sharma, Steffen Schneider, Matthias Bethge, and
  Wieland Brendel.
\newblock Contrastive learning inverts the data generating process, 2021.
\newblock URL \url{https://arxiv.org/abs/2102.08850}.

\end{thebibliography}
\bibliographystyle{plainnat}

\appendix

\section{Appendix}

\subsection{Derivation of the ELBO} \label{elbo_derive}

From equation \ref{eq7}, we want to minimize the KL divergence between the true and approximate posteriors:

\begin{align*}
     &D_{\text{KL}}(q_{\phi}(\mathbf{Z}, \mathcal{G}, \Theta)\, ||\, p(\mathbf{Z}, \mathcal{G}, \Theta \mid \mathcal{D})) \\
     &= - \mathbb{E}_{(\mathbf{Z}, {\mathcal{G}}, \Theta) \sim q_{\phi}(\mathbf{Z}, \mathcal{G}, \Theta)} \biggl[ \log p(\mathcal{D} \mid \mathbf{Z}, {\mathcal{G}}, \Theta) - \log  \frac{ q_{\phi}(\mathcal{G}, \Theta) }{p(\mathcal{G}, \Theta)}  \biggr] + \log p(\mathcal{D}) \\
     &= - \mathbb{E}_{({\mathcal{G}}, \Theta) \sim q_{\phi}(\mathcal{G}, \Theta)} \biggl[ \mathbb{E}_{\mathbf{Z} \sim q_{\phi}(\mathbf{Z} \mid \mathcal{G}, \Theta)} \,[\, \log p(\mathcal{D} \mid \mathbf{Z}) \, ] - \log  \frac{ q_{\phi}(\mathcal{G}, \Theta) }{p(\mathcal{G}, \Theta)}  \biggr] + \log p(\mathcal{D}) \\
     &= - \mathbb{E}_{(P, L, \Sigma) \sim q_{\phi}(P, L, \Sigma)} \biggl[ \mathbb{E}_{\mathbf{Z} \sim q_{\phi}(\mathbf{Z} \mid P, L, \Sigma)} \,[\, \log p(\mathcal{D} \mid \mathbf{Z}) \, ] - \log  \frac{ q_{\phi}(P, L, \Sigma) }{p(P, L, \Sigma)}  \biggr] + \log p(\mathcal{D}) \tag{from \ref{eq8}} \\
     &= - \mathbb{E}_{(L, \Sigma) \sim q_{\phi}(L, \Sigma)} 
     \Biggl[ \mathbb{E}_{P \sim q_{\phi}(P \mid L, \Sigma)} 
     \biggl[ \mathbb{E}_{\mathbf{Z} \sim q_{\phi}(\mathbf{Z} \mid P, L, \Sigma)} 
     \,[\, \log p(\mathcal{D} \mid \mathbf{Z}) \, ] - \log  \frac{ q_{\phi}(P \mid L, \Sigma) }{p(P)} \biggr] \\
     & \hspace{10em}- \log  \frac{ q_{\phi}(L, \Sigma) }{p(L)p( \Sigma)} \Biggr] + \log p(\mathcal{D}) \tag{via the factorization in \ref{eq8}}
\end{align*}

Since the log evidence $\log p(\mathcal{D})$ is a constant, minimizing the KL divergence corresponds to maximizing the following ELBO:

\begin{align*}
    \max \mathbb{E}_{(L, \Sigma) \sim q_{\phi}(L, \Sigma)} 
     \Biggl[ \mathbb{E}_{P \sim q_{\phi}(P \mid L, \Sigma)} 
     \biggl[ \mathbb{E}_{\mathbf{Z} \sim q_{\phi}(\mathbf{Z} \mid P, L, \Sigma)} 
     \,[\, \log p(\mathcal{D} \mid \mathbf{Z}) \, ] - \log  \frac{ q_{\phi}(P \mid L, \Sigma) }{p(P)} \biggr] \\
     - \log  \frac{ q_{\phi}(L, \Sigma) }{p(L)p( \Sigma)} \Biggr] 
\end{align*}

\subsection{Implementation details}

\subsection{Training curves} \label{train_curves}

\begin{figure}[ht]
    \centering
    \includegraphics[width=13cm]{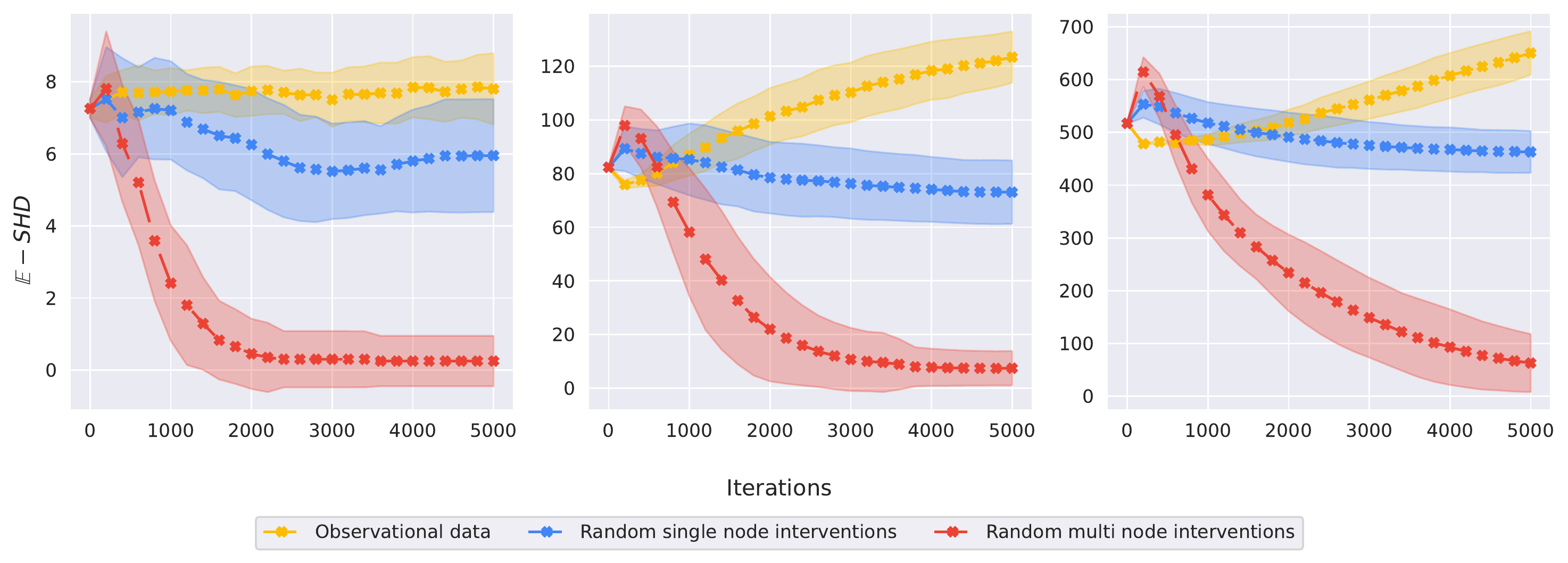}
    \caption{Learning latent SCM parameters given a fixed node ordering for linearly projected causal variables for random ER-1 DAGs with $d = 6, 20, 50$ nodes. The model was trained for 5000 iterations over 3500 data samples out of which 500 were observational points for the single and multi node intervention runs.}
\end{figure}

\begin{figure}[ht]
    \centering
    \includegraphics[width=13cm]{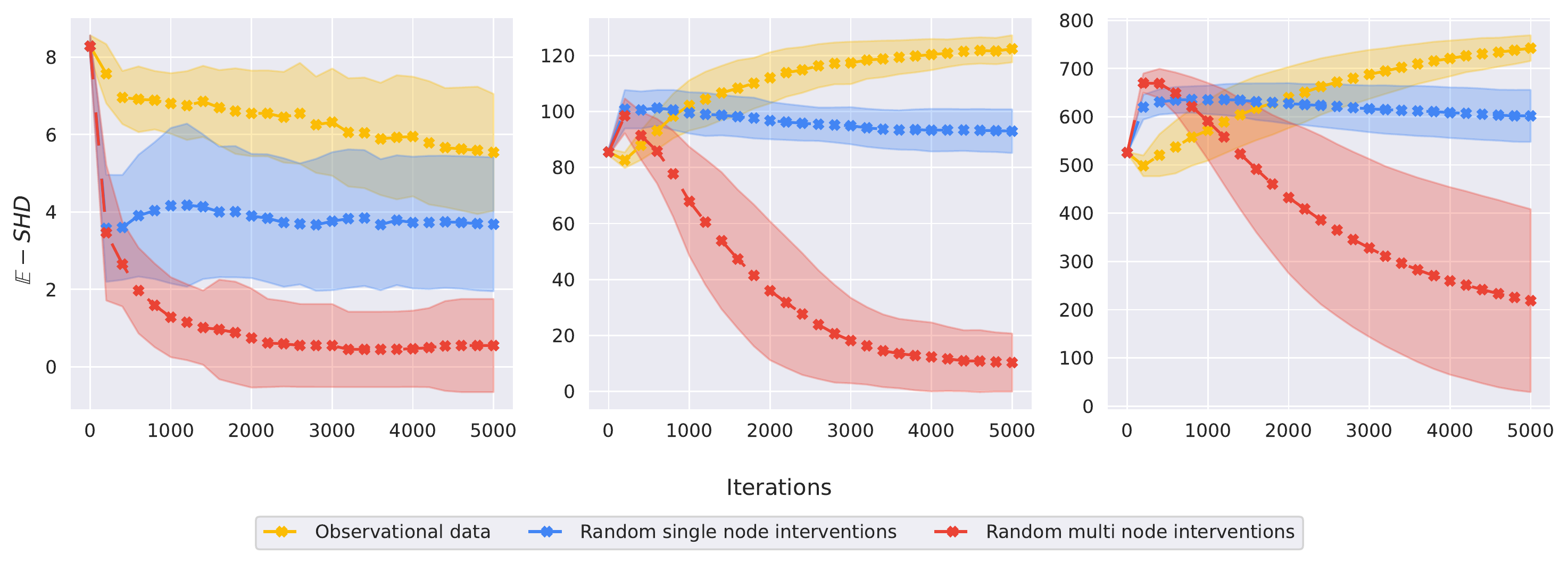}
    \caption{Learning latent SCM parameters given a fixed node ordering for linearly projected causal variables for random ER-2 DAGs with $d = 6, 20, 50$ nodes. The model was trained for 5000 iterations over 3500 data samples out of which 500 were observational points for the single and multi node intervention runs.}
\end{figure}

\begin{figure}[ht]
    \centering
    \includegraphics[width=13cm]{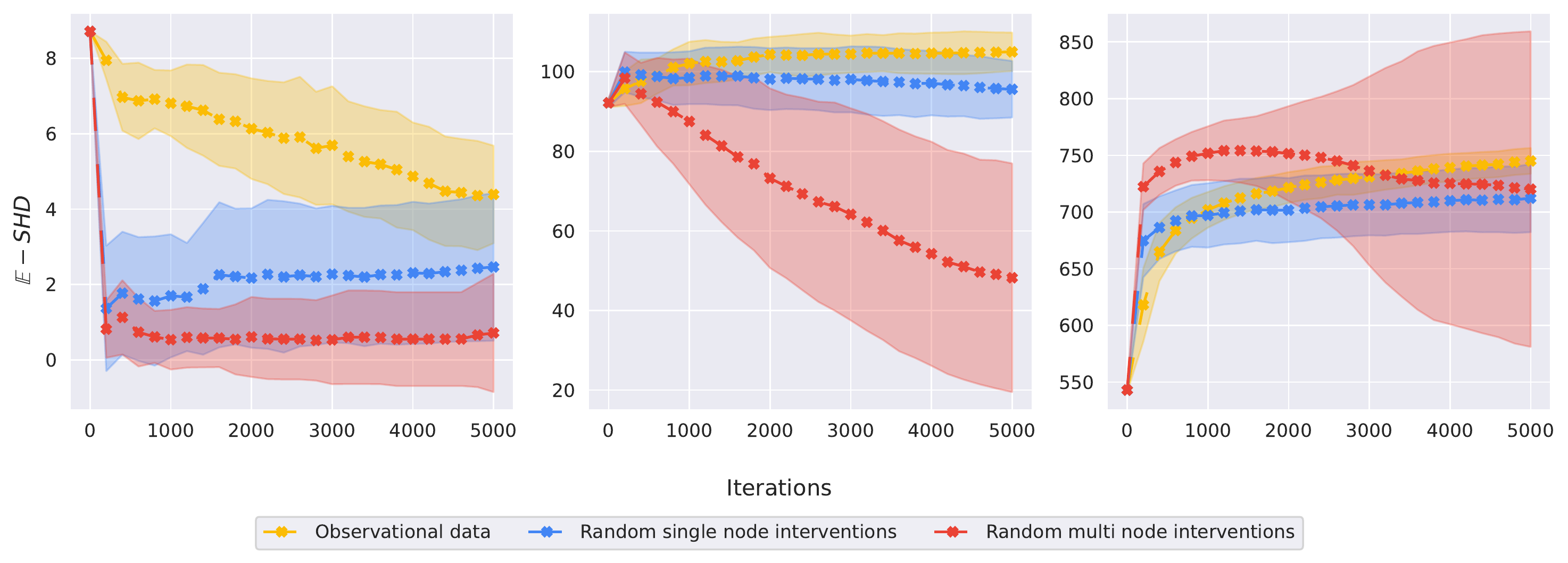}
    \caption{Learning latent SCM parameters given a fixed node ordering for linearly projected causal variables for random ER-4 DAGs with $d = 6, 20, 50$ nodes. The model was trained for 5000 iterations over 3500 data samples out of which 500 were observational points for the single and multi node intervention runs.}
\end{figure}

\subsection{Linear projection experiments}
\begin{figure}[ht]
    \centering
    \includegraphics[width=14cm]{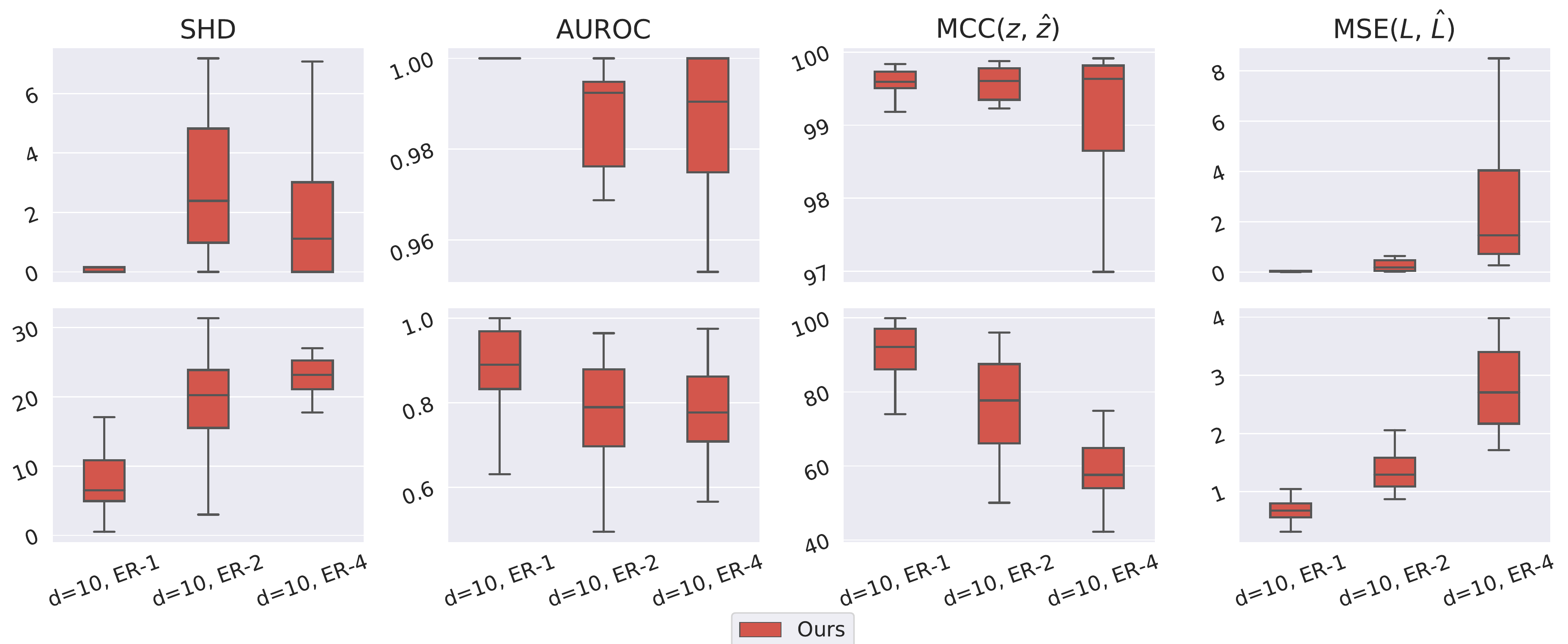}
    \caption{Learning the latent SCM (i) given a node ordering (top) and (ii) over node orderings (bottom) for \textbf{linear} projection of causal variables for $d = 10$ nodes, $\mathrm{D} = 100$ dimensions.}
    \label{d10_linear_dbcd_metrics}
\end{figure}

Details for figure \ref{d10_linear_dbcd_metrics}, $d=10$ nodes: The dataset consists of 500 observational points and 20000 interventional points. To sample the 20000 interventional points, we randomly choose 200 intervention sets, and for each intervention set we sample 100 data points. The model was trained for 8000 epochs to reach convergence.

\begin{figure}[ht]
    \centering
    \includegraphics[width=14cm]{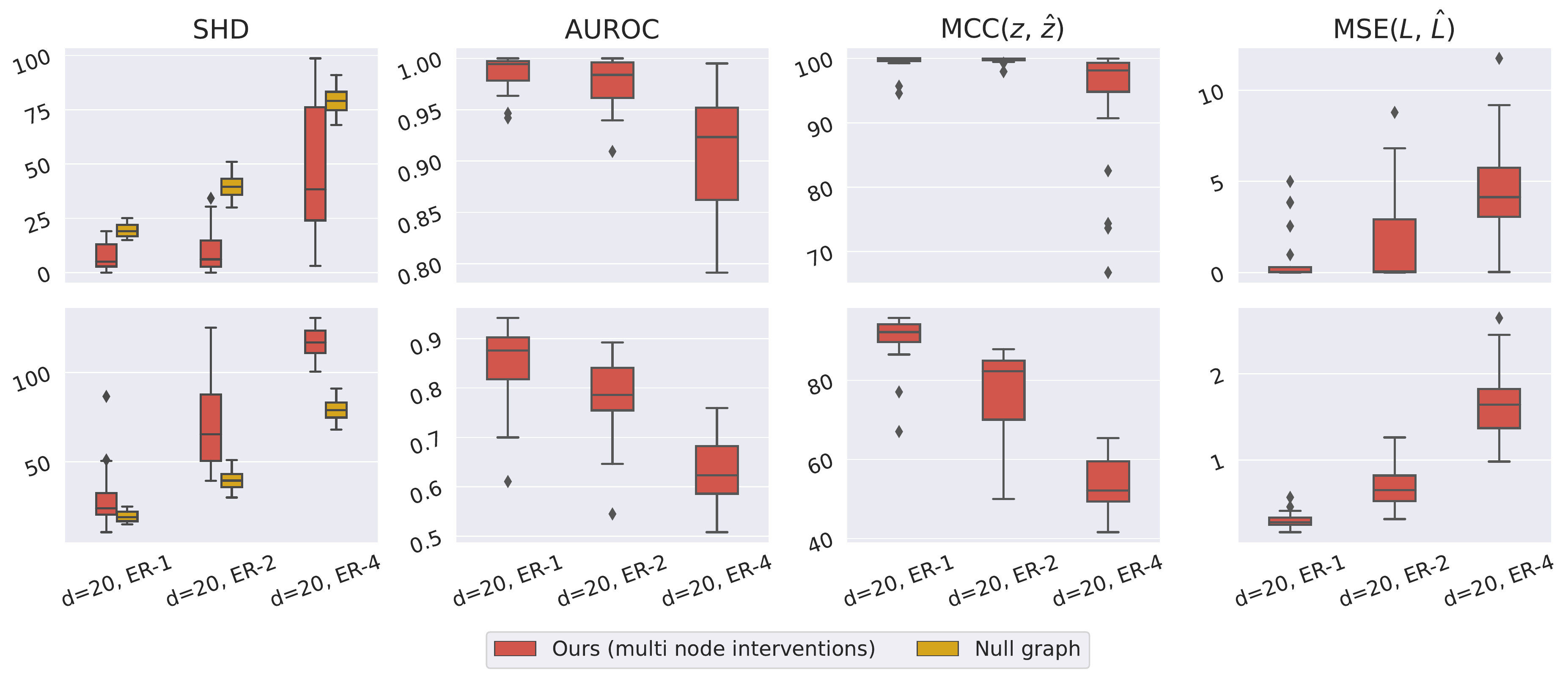}
    \caption{Learning the latent SCM (i) given a node ordering (top) and (ii) over node orderings (bottom) for \textbf{linear} projection of causal variables for $d = 20$ nodes, $\mathrm{D}=100$ dimensions.}
    \label{d20_linear_dbcd_metrics}
\end{figure}

Details for figure \ref{d20_linear_dbcd_metrics}, $d=20$ nodes: The dataset consists of 500 observational points and 20000 interventional points. To sample the 20000 interventional points, we randomly choose 100  intervention sets, and for each intervention set we sample 200 data points. The model was trained for 3000 epochs to reach convergence.

\subsection{Nonlinear projection experiments}

\begin{figure}[ht]
    \centering
    \includegraphics[width=14cm]{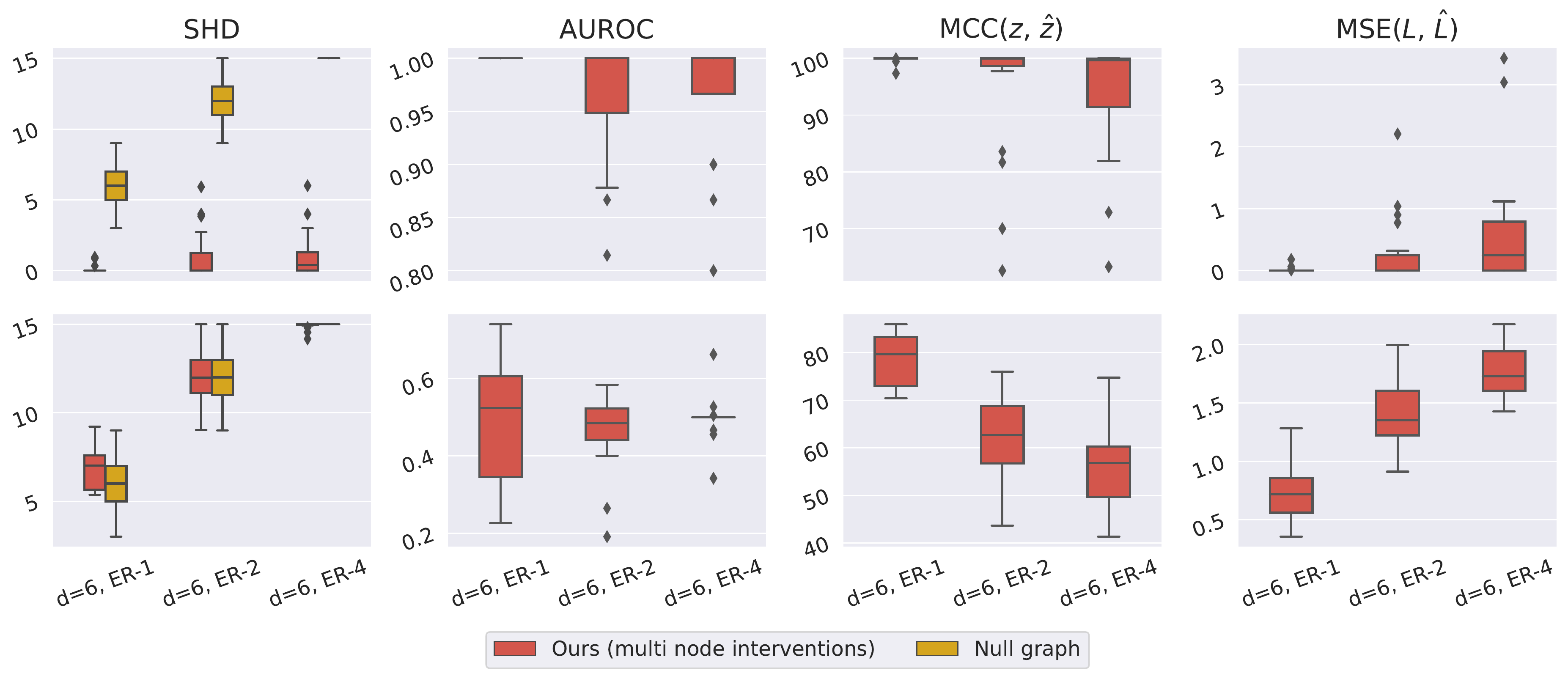}
    \caption{Learning the latent SCM (i) given a node ordering (top) and (ii) over node orderings (bottom) for \textbf{nonlinear} projection of causal variables for $d = 6$ nodes, $\mathrm{D}=100$ dimensions.}
    \label{d6_nonlinear_learn_SCM}
\end{figure}

\begin{figure}[ht] 
    \centering
    \includegraphics[width=14cm]{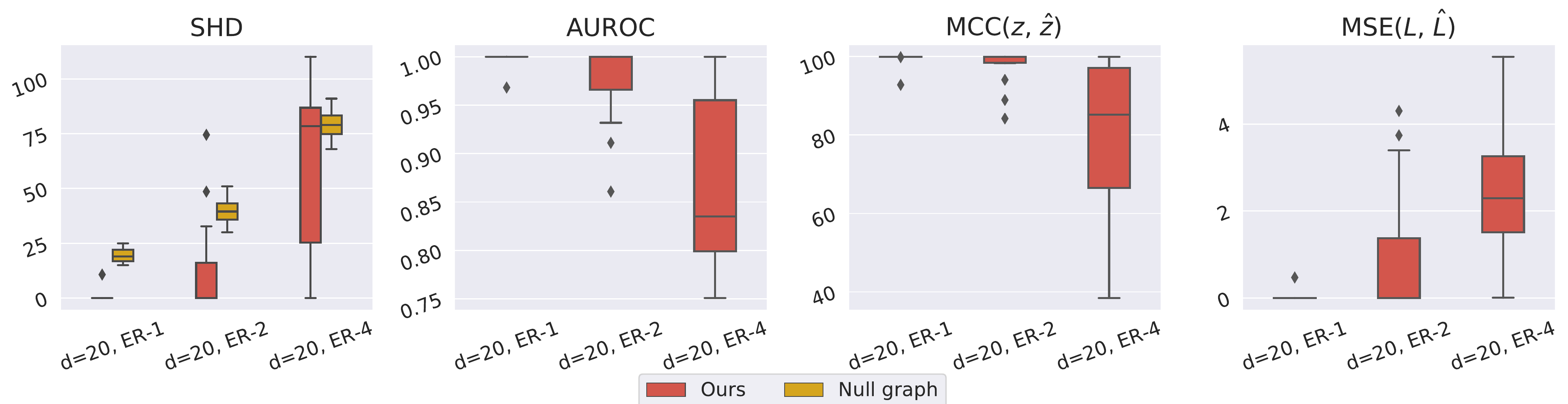}
    \caption{Learning the latent SCM given a node ordering for \textbf{nonlinear} projection of causal variables for $d = 20$ nodes, $\mathrm{D}=100$ dimensions.}
    \label{d20_nonlinear}
\end{figure}

Figure \ref{d6_nonlinear_learn_SCM} contains the results for the latent causal discovery problem with and without learning a permutation. Figure \ref{d20_nonlinear} shows the results for 500 observational samples and 10000 interventional samples with 100 intervention sets and 100 samples per set on $d=20$ nodes and $\mathrm{D}=100$ dimensions.

\subsection{Ablation study: effect of performance with respect to number of intervention types}
\begin{figure}[ht]
    \centering
    \includegraphics[width=14cm]{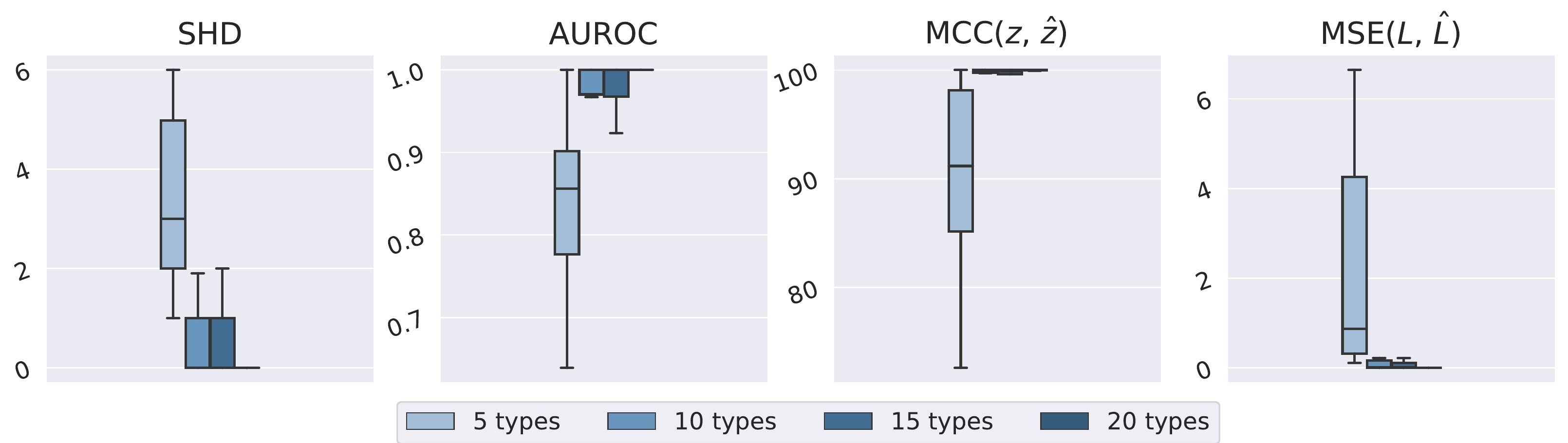}
    \caption{Effect of number of intervention types seen versus on the performance of learning the latent SCM given a node ordering for $d = 5$ nodes, $\mathrm{D}=100$ dimensions.}
    \label{d5_vs_interv_types_linear_dbcd}
\end{figure}

\begin{figure}[ht]
    \centering
    \includegraphics[width=14cm]{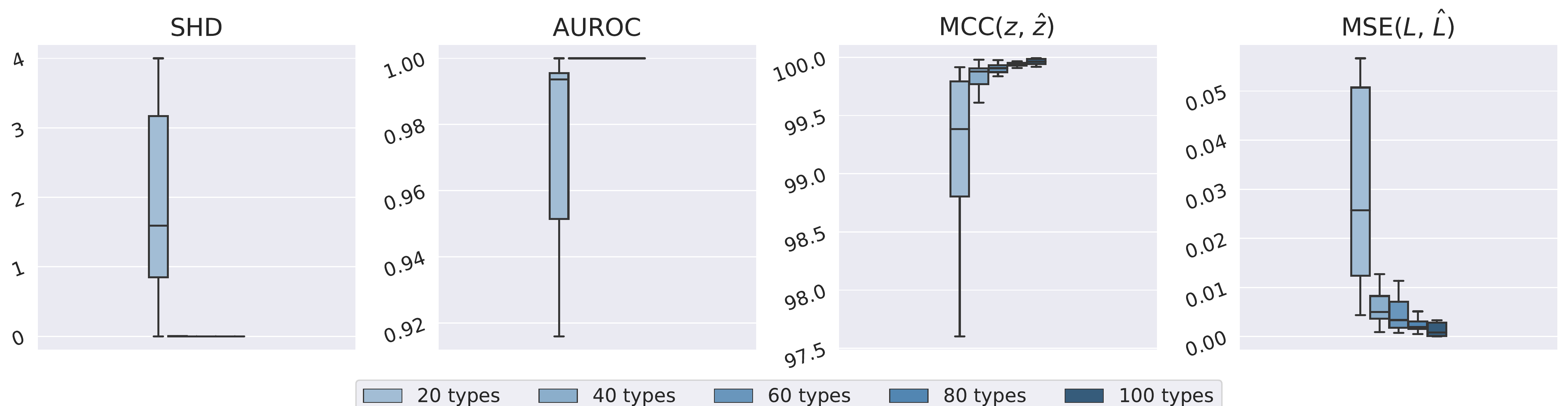}
    \caption{Effect of number of intervention types seen versus on the performance of learning the latent SCM given a node ordering for $d = 10$ nodes, $\mathrm{D}=100$ dimensions.}
    \label{d10_vs_interv_types_linear_dbcd}
\end{figure}

\begin{figure}[ht]
    \centering
    \includegraphics[width=14cm]{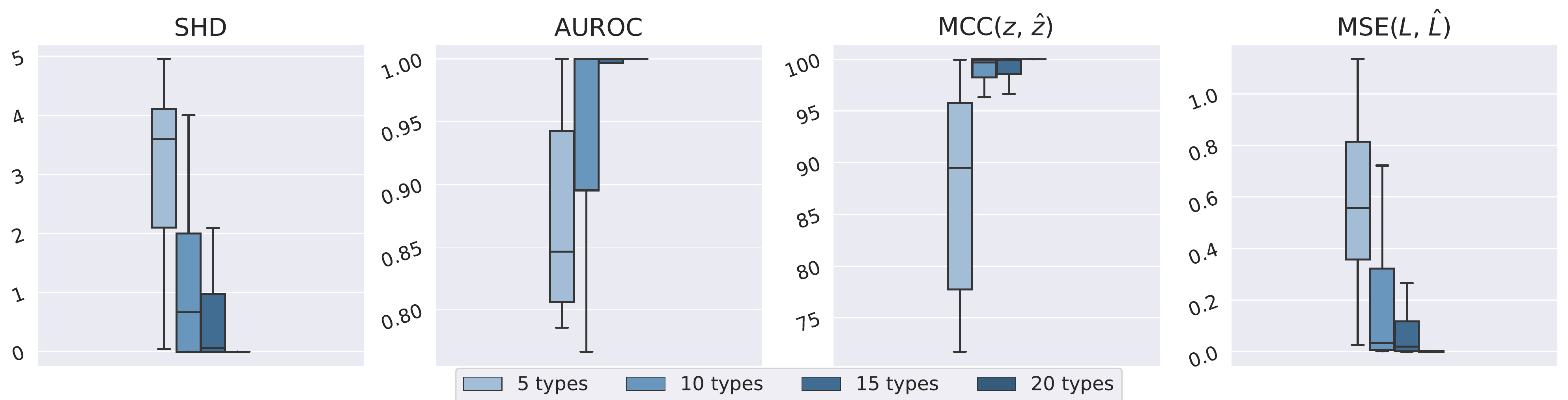}
    \caption{Ablation: effect of number of intervention types seen versus on the performance of learning the latent SCM given a node ordering for $d = 5$ nodes, $\mathrm{D}=100$ dimensions.}
    \label{d5_vs_interv_types_nonlinear_dbcd}
\end{figure}

\begin{figure}[ht]
    \centering
    \includegraphics[width=14cm]{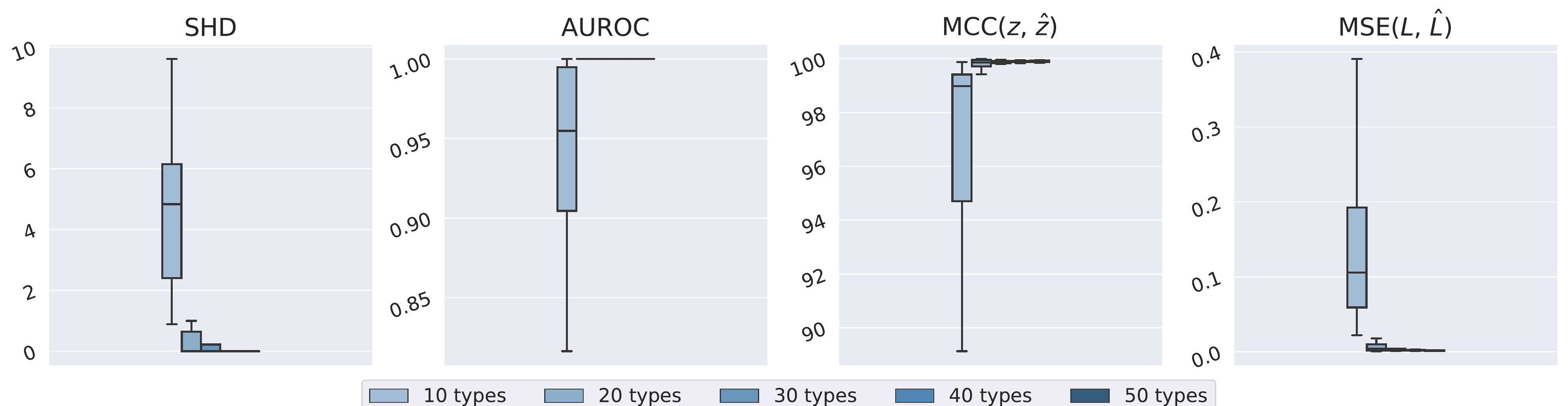}
    \caption{Ablation: effect of number of intervention types seen versus on the performance of learning the latent SCM given a node ordering for $d = 10$ nodes, $\mathrm{D}=100$ dimensions.}
    \label{d10_vs_interv_types_nonlinear_dbcd}
\end{figure}

Here, we study how the performance and recovery of the latent SCM is affected with respect to the number of intervention types in the dataset. The number of intervention types refers to different combinations of nodes we perform interventions on. For all experiments in this subsection, we use 100 interventional samples per type of intervention. Figure \ref{d5_vs_interv_types_linear_dbcd} and \ref{d10_vs_interv_types_linear_dbcd} show results on vector data where the high-dimensional vector is a linear projection of the causal variables.
Figure \ref{d5_vs_interv_types_nonlinear_dbcd} and \ref{d10_vs_interv_types_nonlinear_dbcd} summarize results on vector data where the high-dimensional vector is a nonlinear projection of the causal variables.

\begin{table}[ht]
\centering
\caption{Average experiment runtimes}
\begin{tabular}{ |c|c|c|c|c|c| } 
\hline
Nodes ($d$) & Dataset size & Experiment & Projection & Steps & Avg. runtime (min) \\
\hline
5 & 2500 & Vector data & Linear (fixed ordering) & 5000 & 12 \\ 
10 & 10500 & Vector data & Linear (fixed ordering) & 3000 & 25 \\ 
10 & 10500 & Vector data & Linear (learned ordering) & 3000 & 90 \\ 
20 & 80500 & Vector data & Linear (fixed ordering) & 3000 & 150 \\ 
20 & 20500 & Vector data & Linear (learned ordering) & 3000 & 360 \\
20 & 80500 & Vector data & Linear (learned ordering) & 8000 & 540 \\
10 & 5500 & Vector data & Nonlinear & 2000 & 12 \\
5 & 2500 & Vector data & Nonlinear & 5000 & 15 \\
20 & 10500 & Vector data & Nonlinear & 10000 & 100 \\
5 & 2500 & Image data & Nonlinear & 2000 & 40 \\
5 & 5500 & Image data & Nonlinear & 2000 & 75 \\
10 & 2500 & Image data & Nonlinear & 2000 & 50 \\
10 & 5500 & Image data & Nonlinear & 2000 & 80 \\
\hline
\end{tabular}
\label{table:1}
\end{table}

\subsection{Complete evaluation} \label{detail_eval}
\begin{figure}
    \centering
    \includegraphics[width=12cm]{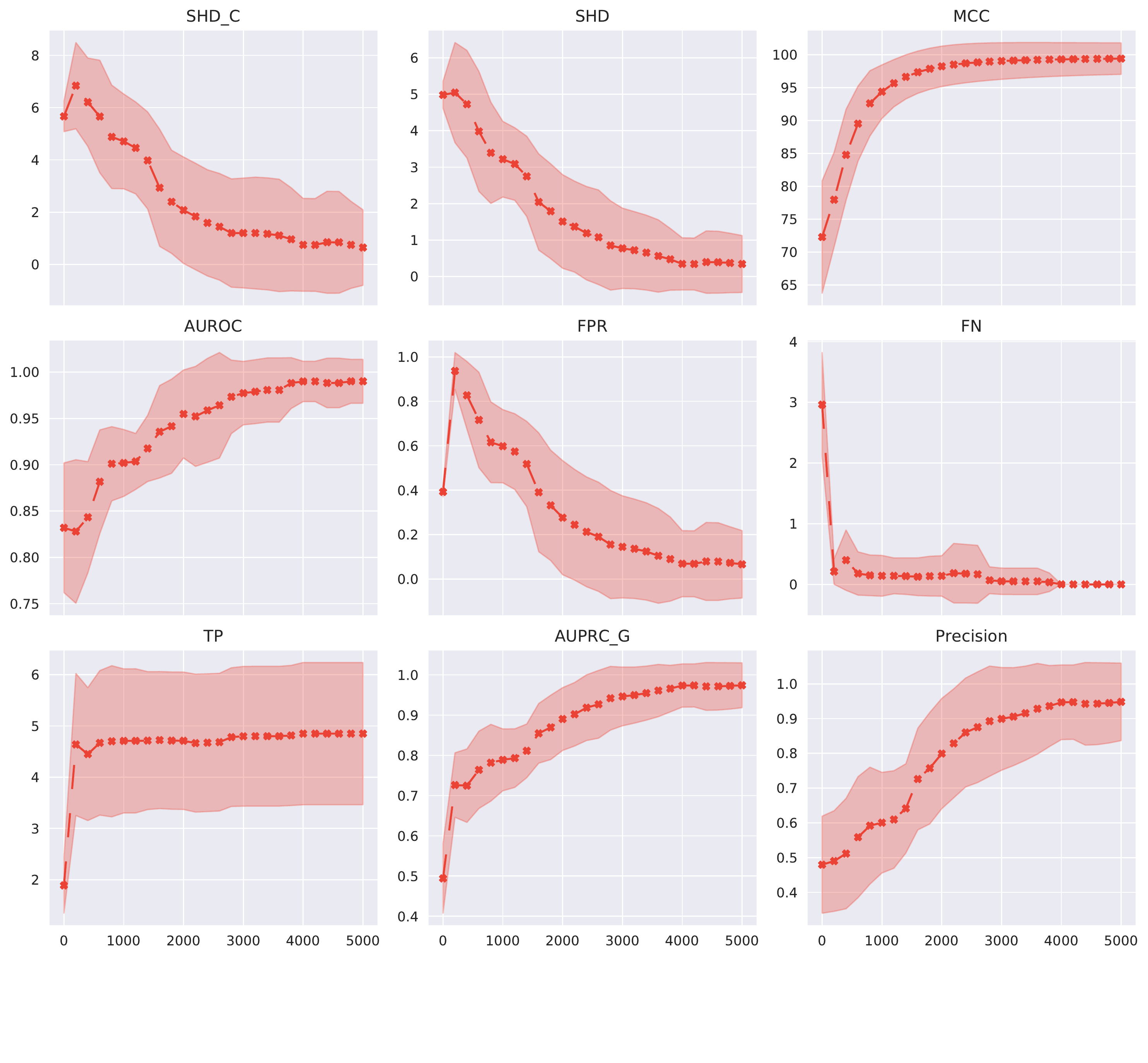}
    \vspace{-1cm}
    \includegraphics[width=12cm]{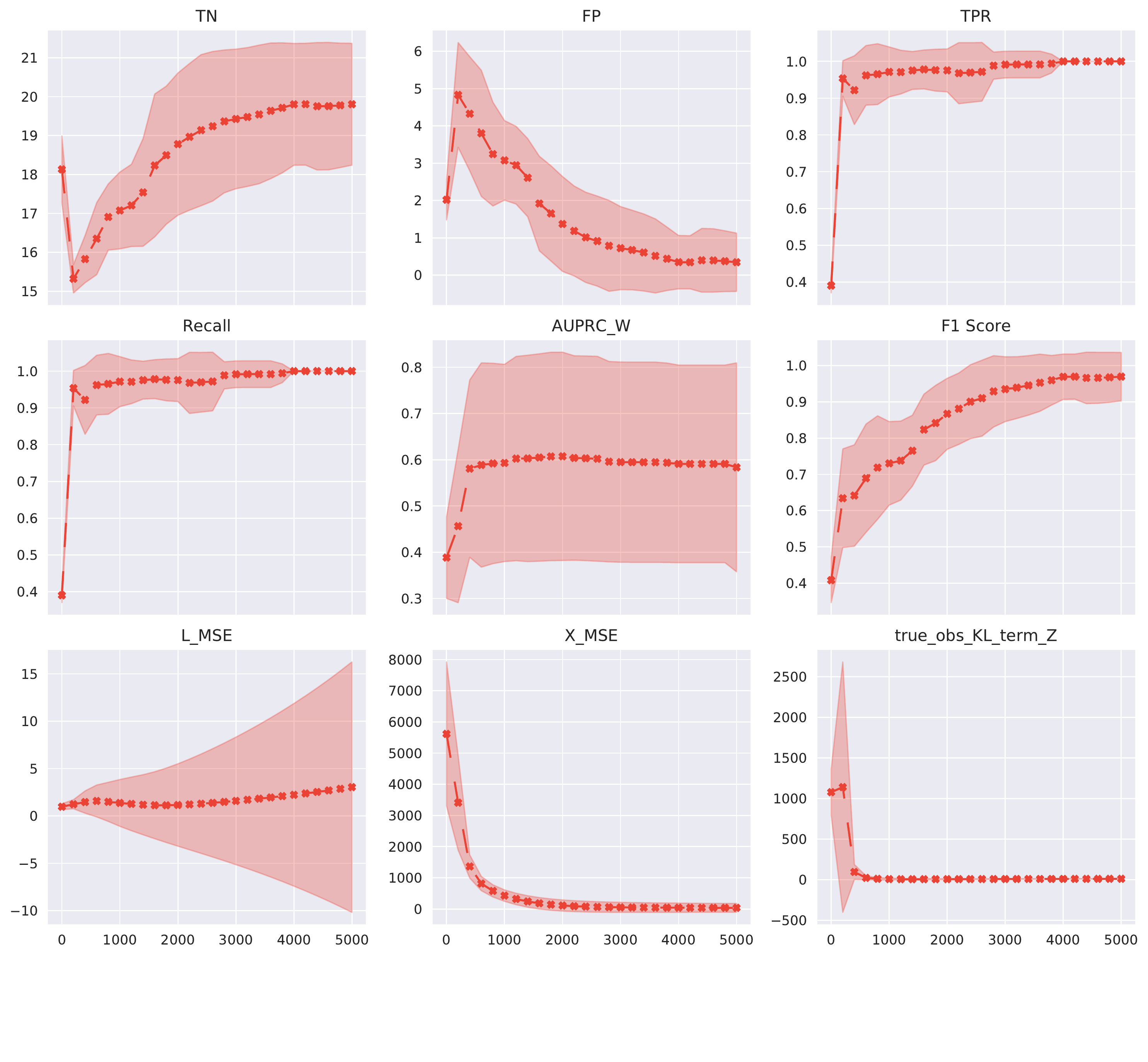}
    \caption{Training curves: Metric versus number of iterations for $d=5$ nodes linearly projected to $\mathrm{D}=100$ dimensions, with 20 intervention sets, 100 interventional samples per intervention set, and 500 observational samples.}
    \label{all_metrics_d5_linear_dbcd}
\end{figure}

For the experiments already presented in the main text, this section contains additional a more comprehensive evaluation on the following metrics: 

\begin{itemize}[noitemsep]
    \item \textbf{SHD\_C}: The expected CPDAG SHD between the GT and predicted DAG's skeletons.
    \item \textbf{SHD}: The expected SHD between the GT and predicted DAG.
    \item \textbf{MCC}: The expected Mean Correlation Coefficient between the predicted and true causal variables. 
    \item \textbf{AUROC} between predicted and true graph structure.
    \item \textbf{FPR}: False positive rate
    \item \textbf{FN}: False negative
    \item \textbf{TP}: True positive
    \item \textbf{AUPRC\_G}: Area under precision-recall curves
    \item \textbf{Precision} of the graph structure prediction 
    \item \textbf{TN}: True negatives
    \item \textbf{FP}: False positives
    \item \textbf{TPR}: True positive rate
    \item \textbf{Recall} of the graph structure prediction 
    \item \textbf{AUPRC\_W}:
    \item \textbf{F1 Score} $=2*Precision*Recall/(Precision + Recall)$
    \item \textbf{L\_MSE}: Mean squared error between predicted and true edge weights.
    \item \textbf{X\_MSE}: Mean squared reconstruction error over the high dimensional (low-level) data
    \item \textbf{true\_obs\_KL\_term\_Z}: The KL divergence between the predicted and GT observational joint distributions. 
\end{itemize}

Figure \ref{all_metrics_d5_linear_dbcd} shows the complete evaluation on 18 different metrics for $d=5$ nodes. The dataset consists of 500 observational points and 2000 interventional points. To sample the 2000 interventional points, we randomly choose 20 intervention sets, and for each intervention set we sample 100 data points with random intervention values.

\subsection{Additional related work} \label{more_rel_work}

\textbf{Causal discovery and structure learning}: \cite{vcn} casts the Bayesian structure learning problem as an autoregressive one by sequentially predicting edges, in hopes of capturing the potentially multi-modal posterior. \cite{daggfn} uses Generative Flow Networks, or GFlowNets~\citep{bengio2021flow}, a new class of probabilistic methods that lies at the intersection of reinforcement learning and variational inference. The work uses the transitive closure property ensuring that the action space is constrained to actions that do not introduce cycles. \cite{ges} proposes a greedy search algorithm, but does not scale to large number of nodes. \cite{trust} leverages sum product networks to perform exact Bayesian structure learning.~\cite{hagele2022bacadi} extends the framework of~\cite{dibs} to perform Bayesian causal discovery in a setting where interventions are unknown. \cite{LaHME} is in a setting where the edges exist not just between latent causal variables but with high-dimensional variables in the dataset as well. Other efforts include \citep{shimizu2011directlingam, rctt, dag_gnn, ghoshal2018learning, ng2020role, ircm}.

\end{document}